\pdfoutput=1

\documentclass[11pt]{article}

\usepackage[preprint]{acl}

\usepackage{times}
\usepackage{latexsym}

\usepackage{amsmath}
\usepackage{amssymb}
\usepackage{subcaption}
\usepackage{booktabs}
\usepackage{algpseudocode}

\usepackage[T1]{fontenc}

\usepackage[utf8]{inputenc}

\usepackage{microtype}

\usepackage{inconsolata}

\usepackage{graphicx}

\title{Reasoning Circuits in Language Models:\\ A Mechanistic Interpretation of Syllogistic Inference}

\author{\textbf{Geonhee Kim$^1$, Marco Valentino$^{2,3}$, Andr\'e Freitas$^{1,2,4}$} \\
$^{1}$Department of Computer Science, University of Manchester, UK \\ 
$^{2}$Idiap Research Institute, Switzerland \\
$^{3}$School of Computer Science, University of Sheffield, UK \\ 
$^{4}$National Biomarker Centre, CRUK-MI, University of Manchester, UK\\
\tt{kimgeonhee317@gmail.com},
\tt{ac4mv@sheffield.ac.uk}, \tt{andre.freitas@idiap.ch}}

\begin{document}
\maketitle
\begin{abstract}
Recent studies on reasoning in language models (LMs) have sparked a debate on whether they can learn systematic inferential principles or merely exploit superficial patterns in the training data. To understand and uncover the mechanisms adopted for formal reasoning in LMs, this paper presents a mechanistic interpretation of syllogistic inference. Specifically, we present a methodology for circuit discovery aimed at interpreting content-independent and formal reasoning mechanisms. Through two distinct intervention methods, we uncover a sufficient and necessary circuit involving middle-term suppression that elucidates how LMs transfer information to derive valid conclusions from premises.
Furthermore, we investigate how belief biases manifest in syllogistic inference, finding evidence of partial contamination from additional attention heads responsible for encoding commonsense and contextualized knowledge. Finally, we explore the generalization of the discovered mechanisms across various syllogistic schemes, model sizes and architectures. The identified circuit is sufficient and necessary for syllogistic schemes on which the models achieve high accuracy ($\geq$ 60\%), with compatible activation patterns across models of different families. Overall, our findings suggest that LMs learn transferable content-independent reasoning mechanisms, but that, at the same time, such mechanisms do not involve generalizable and abstract logical primitives, being susceptible to contamination by the same world knowledge acquired during pre-training.
\end{abstract}

\section{Introduction}

\begin{figure}[t!]
\centering
\includegraphics[width=\linewidth]{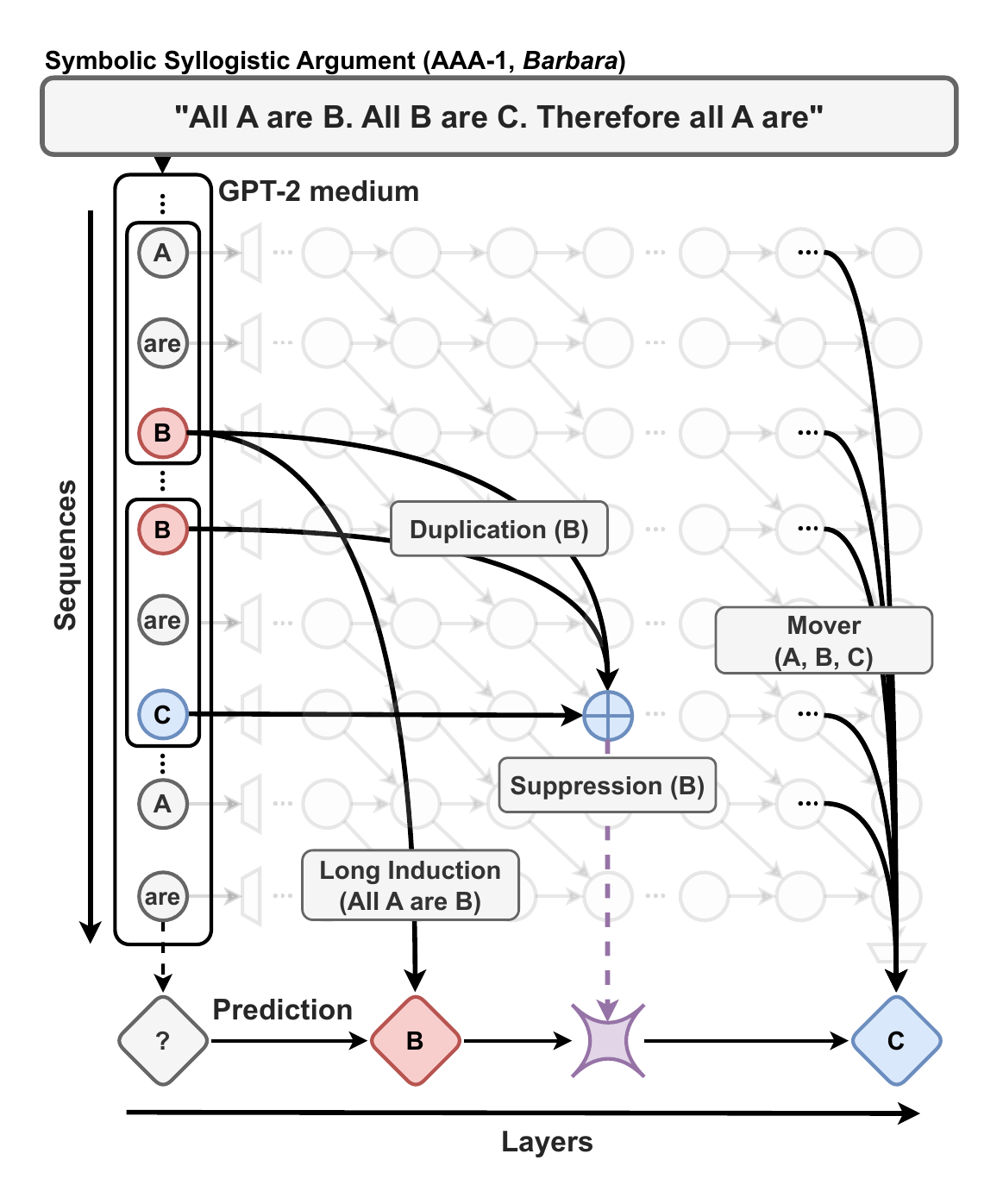}
\caption{Conceptual representation of the circuit for processing symbolic syllogisms: (Long Induction) Early layers exhibit biases towards the wrong conclusion due to long-range repetition of the first premise ``All A are B''. (Duplication) Induction heads aggregate information about duplicated middle terms. (Suppression) The model inhibits middle-term information (i.e., `B'), suppressing the long induction mechanism. (Mover) Token-specific information is propagated to the last token position. The process culminates in the prediction shift from `B' to the correct token, `C.'}
\label{fig:circuit_intro}
\end{figure}

Language models (LMs) have led to remarkable results across various natural language processing tasks \citep{Radford2018, Radford2019, Brown2020, Achiam2023, Jason2022, Bubeck2023}. This success has catalyzed research interest in systematically exploring the reasoning capabilities emerging during pre-training \citep{Clark2020}.
Recent findings suggest that logical and formal reasoning abilities may emerge in large-scale models \citep{Rae2021, Kojima2022, Wei2022} or through transfer learning on specialized datasets \citep{Betz2021}. However, ongoing debates question whether these models apply systematic inference rules or reuse superficial patterns learned during pre-training \citep{talmor2020, Kassner2020, Wu2024}. This controversy underscores the need for a deeper understanding of the low-level logical inference mechanisms in LMs \citep{Rozanova2024, Rozanova2023,rozanova-etal-2022-decomposing,yanaka-etal-2020-neural}.

To improve our understanding of the internal mechanisms, this paper focuses on mechanistic interpretability \citep{Olah2020, Nanda2023}, aiming to discover the core circuit responsible for syllogistic reasoning. In particular, we focus on categorical syllogisms with universal affirmative quantifiers (i.e., AAA-1, \textit{Barbara}) motivated by two key factors. First, as observed in natural logic studies \citep{MacCartney2007}, this form of syllogistic reasoning is prevalent in everyday language. Therefore, it is likely that LMs are exposed to such reasoning schema during pre-training. Second, AAA-1 is a form of unconditionally valid syllogism independent of the premises' truth condition \citep{Holyoak2005}. This characteristic offers a deterministic and scalable task design, other than allowing us to investigate the disentanglement between reasoning and knowledge representation \cite{bertolazzi-etal-2024-systematic,wysocka2024syllobio,lampinen2024language,valentino2025mitigating}. 

Through mechanistic interpretability techniques such as Activation Patching \citep{Meng2022} and embedding space analysis (i.e., Logit Lens)~\citep{Nostalgebraist2020, Geva2022, Dar2023}, we investigate the following main research questions: \emph{RQ1: How is the content-independent syllogistic reasoning mechanism internalized in LMs during pre-training?}; \emph{RQ2: Are content-independent mechanisms disentangled from specific world knowledge and belief biases?}; \emph{RQ3: Does the core reasoning mechanism generalize across syllogistic schemes, different model sizes, and architectures?}

To answer these questions, we present a methodology that consists of 3 main stages. First, we define a syllogistic completion task designed to assess the model's ability to predict valid conclusions from premises and facilitate the construction of test sets for circuit analysis. Second, we implement a circuit discovery pipeline on the syllogistic schema instantiated only with symbolic variables (Table \ref{tab:example_1}, Symbolic) to identify the core sub-components responsible for content-independent reasoning. We conduct this analysis under two intervention methods: \emph{middle-term corruption} and \emph{all-term corruption}, aiming to identify latent transitive reasoning mechanisms and term-related information flow.
Third, we investigate the generalization of the identified circuit on concrete schemes instantiated with commonsense knowledge to identify potential belief biases and explore how the internal behavior varies with different schemes and model sizes.

We present the following overall conclusions:

1. The circuit analysis reveals that LMs develop specific inference mechanisms during pre-training, finding evidence supporting a three-stage mechanism for syllogistic reasoning: (1) naive recitation of the first premise; (2) suppression of duplicated middle-term information; and (3) mediation towards the correct output through the interplay of mover attention heads (see Figure~\ref{fig:circuit_intro}).

2. Further experiments on circuit transferability demonstrate that the identified mechanism is still necessary for reasoning on syllogistic schemes instantiated with commonsense knowledge. However, a deeper analysis suggests that specific belief biases acquired during pre-training might contaminate the content-independent circuit mechanism with additional attention heads responsible for encoding contextualized world knowledge.

3. We found that the identified circuit is sufficient and necessary for all the unconditionally valid syllogistic schemes in which the model achieves high downstream accuracy ($\geq$ 60\%) (see Appendix~\ref{app:appendix_A} for the list of schemes). This result suggests that LMs learn reasoning mechanisms that are transferable across different schemes.

4.  The intervention results on models with different architectures and sizes (i.e., GPT-2~\citep{Radford2019}, Pythia~\citep{pythia}, Llama~\citep{llama3}, Qwen~\citep{qwen2.5}) show similar suppression mechanism patterns and information flow. However, we found evidence that the contribution of attention heads becomes more complex with model sizes, further supporting the hypothesis of increasing contamination from external world knowledge.

\begin{table}[t!]
\small
\centering
\begin{tabular}{p{4cm}p{3cm}}
\toprule
\textbf{Premises ($\mathcal{P}_1, \mathcal{P}_2$)} & \textbf{Conclusion ($\mathcal{C}$)} \\
\midrule
\multicolumn{2}{l}{\textbf{Symbolic}} \\
All A are B. \newline All B are C. & Therefore,\newline all A are \underline{\textbf{C}}. \\ 
\midrule
\multicolumn{2}{l}{\textbf{Belief-consistent (True premises)}} \\
All men are humans. \newline All humans are mortal. & Therefore,\newline all men are \underline{\textbf{mortal}}. \\ 
\midrule
\multicolumn{2}{l}{\textbf{Belief-inconsistent (False premises)}} \\
All pilots are people. \newline All people are blond. & Therefore,\newline all pilots are \underline{\textbf{blond}}. \\ 
\bottomrule
\end{tabular}
\caption{Examples of a syllogistic schema (i.e., AAA-1, Barbara). The logical validity of a conclusion is only a function of the reasoning schema, being independent of the specific variables or truth condition of the premises.}
\label{tab:example_1}
\end{table}

\begin{figure*}[t!]
\centering
\includegraphics[width=\linewidth]{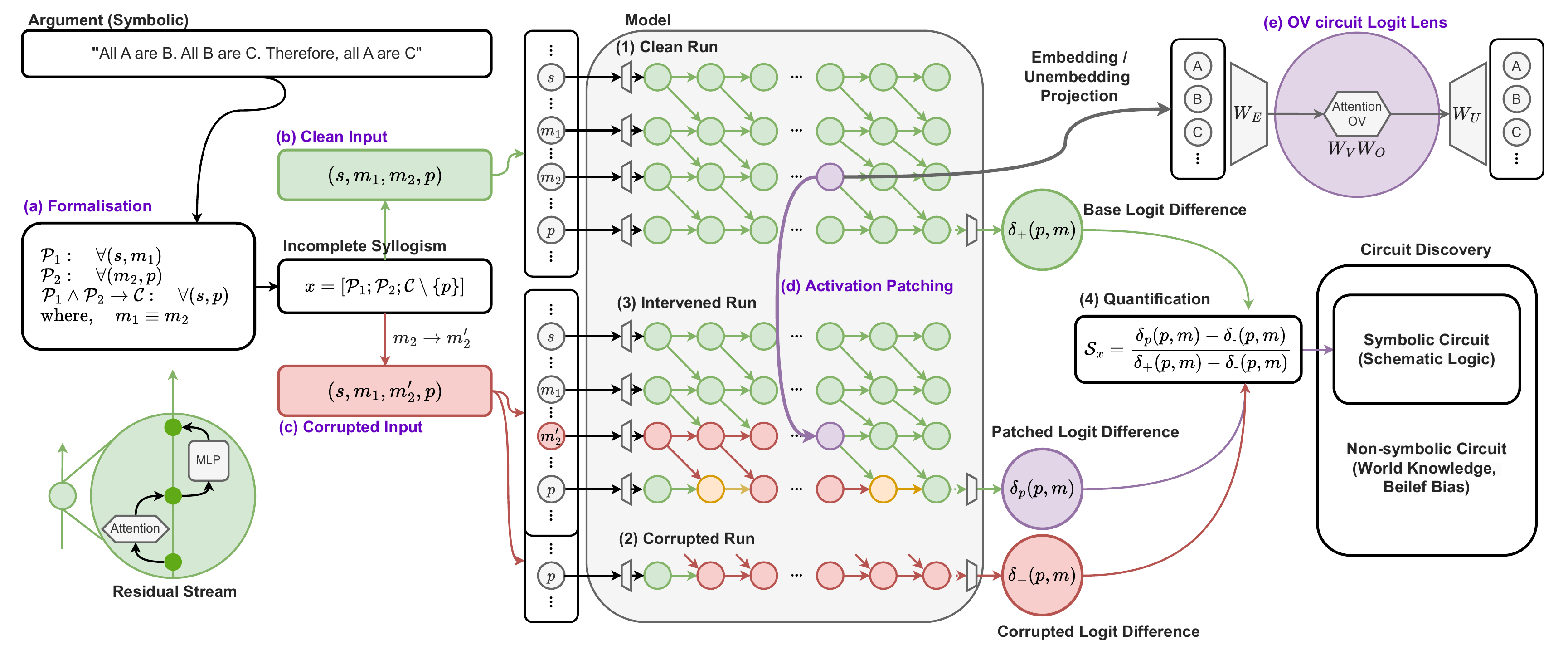}
\caption{
The conceptual pipeline of the circuit discovery methodology. Figures (a)--(e) illustrate the stages in processing a syllogistic argument: formalization (a), construction of clean (b) and corrupted inputs (c), activation patching (d) and embedding space analysis (i.e., Logit Lens) (e). The uncovered symbolic circuit is evaluated to determine whether the core reasoning schema operates independently from world knowledge or belief bias.
}

\label{fig:methodology}
\end{figure*}

Our study is conducted using TransformerLens~\citep{Nanda2022} on an Nvidia A100 GPU with 80GB of memory. The dataset and code to reproduce our experiments are available online to encourage future work in the field\footnote{\url{https://github.com/neuro-symbolic-ai/Mechanistic-Interpretation-Syllogism}}.

\section{Methodology}

Our main research objective is to discover and interpret the core mechanisms adopted by auto-regressive language models (LMs) when performing content-independent syllogistic reasoning. To this end, we present a methodology that consists of 3 main stages. First, we define a syllogistic completion task that can be instrumental for our analysis. Second, we leverage the syllogistic completion task to implement a circuit discovery pipeline on a syllogistic schema instantiated only with symbolic variables (Table \ref{tab:example_1}, Symbolic). Third, we investigate the generalization of the identified circuit on concrete schemes instantiated with commonsense knowledge and explore how the internal behavior varies with different schemes and models.

\subsection{Syllogism Completion Task}
\label{sec:syllogism_completion}

We design a syllogism completion task for assessing the reasoning abilities of LMs, building upon established approaches in the literature \citep{Betz2021, Wu2023}. In particular, we focus on categorical syllogisms with universal affirmative quantifiers (i.e., AAA-1, \textit{Barbara}) because of their frequency in natural language and their content-independent reasoning property. 
The syllogistic argument is typically composed of two premises ($\mathcal{P}_1, \mathcal{P}_2$) and a conclusion ($\mathcal{C}$), with $s$ and $p$ denoting the subject and predicate terms in the conclusion (e.g. \emph{men} and \emph{mortal}), and $m_1$ and $m_2$ (e.g., \emph{humans}) denoting the middle terms in the two premises, with $m_1 \equiv m_2$ in the case of the AAA-1 syllogism.

To evaluate syllogistic reasoning in LMs, we formalize the completion task as language modeling, removing the final token $p$ from the conclusion (e.g., \textit{All men are humans. All humans are mortal. Therefore, all men are}), and comparing the probability assigned by the LM to $p$ (e.g., \textit{mortal}) and the middle term $m$ (e.g., \textit{humans}). In general, an LM is successful in the completion of a task if the following condition applies:
\begin{equation*}
P(p \mid [\mathcal{P}_1; \mathcal{P}_2; \mathcal{C} \setminus \{p\}]) > P(m \mid [\mathcal{P}_1; \mathcal{P}_2; \mathcal{C} \setminus \{p\}])
\end{equation*}

In our experiments, we measure the logit difference $\delta$ between the tokens $p$ and $m$, defined as $\delta(p, m) = \text{logit}(p) - \text{logit}(m)$, which approximates the log ratio of the conditional probabilities:
\begin{equation*}
\delta(p, m) \approx \log{\left(\frac{P(p \mid [\mathcal{P}_1 ; \mathcal{P}_2 ; \mathcal{C} \setminus \{p\}])} {P(m \mid [\mathcal{P}_1 ; \mathcal{P}_2 ; \mathcal{C} \setminus \{p\}])} \right)}
\end{equation*}

\paragraph{Dataset Construction.} We conduct experiments using two distinct datasets, symbolic and non-symbolic, to derive comparative implications for model reasoning capabilities. The symbolic dataset is constructed by randomly sampling triples (e.g., A, B, and C) from the set of 26 uppercase letters of the English alphabet. On the other side, the non-symbolic dataset is constructed by replacing the letters with words while preserving the syllogistic schema. Here, we generate two different non-symbolic sets: a belief-consistent set containing true premises and a belief-inconsistent set containing false premises to address RQ2 (see Table \ref{tab:example_1}). To guarantee the truth condition of the premises, we leverage GenericsKB \citep{Bhakthavatsalam2020}, a knowledge base containing statements about commonsense knowledge. The detailed generation process is in Appendix~\ref{app:appendix_B}.

\subsection{Circuit Discovery} 
Our main objective is to find a circuit $\mathcal{C}$ for syllogistic reasoning in a language model $\mathcal{M}$. A circuit $\mathcal{C}$ can be defined as a subset of the original model $\mathcal{M}$ that is both sufficient and necessary for achieving the original model performance on the syllogistic completion task.
In order to identify a circuit, we employ activation patching 
together with circuit ablation~\citep{Meng2022, Vig2020}. 

\paragraph{Activation Patching.} This technique involves modifying the activation of targeted components and observing the resulting changes. Our study primarily examines the activation of residual streams and multi-head self-attention at the sequence level to trace the term information flow. Activation patching includes three model runs (Clean, Corrupted and Intervened) alongside a quantification process to measure the effect of the interventions (see Figure~\ref{fig:methodology}(1)--(4)). 
Given a masked syllogistic input $x = [\mathcal{P}_1 ; \mathcal{P}_2 ; \mathcal{C} \setminus \{p\}]$ as $(s, m_1, m_2, p)$, which can be read as ``All $s$ are $m_1$. All $m_2$ are $p$. Therefore, all $s$ are [mask]'', and its correct completion $y = p$, the activation patching technique consists of the following steps:

\paragraph{(1) Clean Run.} For the target syllogistic input $(s, m_1, m_2, p)$, we record the baseline logit difference, $\delta_{+}(p, m)$ from the forward pass output of the model (Figure~\ref{fig:methodology}(1)).

\paragraph{(2) Corrupted Run.} We re-run the model on a corrupted input after applying a specific intervention (e.g., changing the middle term $m_2$) and record the logit difference $\delta_{-}(p, m)$ (Figure~\ref{fig:methodology}(3)).

\paragraph{(3) Intervened Run.} We run the model with the corrupted inputs again and replace the corrupted activations with those from the clean runs to compute the response from the remaining components and measure the causal impact of the intervention. Here, we measure the adjusted logit difference $\delta_{p}(p, m)$ (Figure~\ref{fig:methodology}(2)).

\paragraph{(4) Quantification.} We quantify the causal impact of each intervention using a patching score $\mathcal{S}$ following~\citep{Heimersheim2024} as shown in Figure~\ref{fig:methodology}(4). This score is further normalized to $[-1, 1]$.

We complement Activation Patching with known methods for analysing hidden activations in transformers, including Logit Lens~\citep{Nostalgebraist2020} with an input-agnostic approach~\citep{Dar2023, Hanna2024}. Additional technical details can be found in Appendix~\ref{app:appendix_C}.

\begin{figure*}[h!]
\centering
\includegraphics[width=\linewidth]{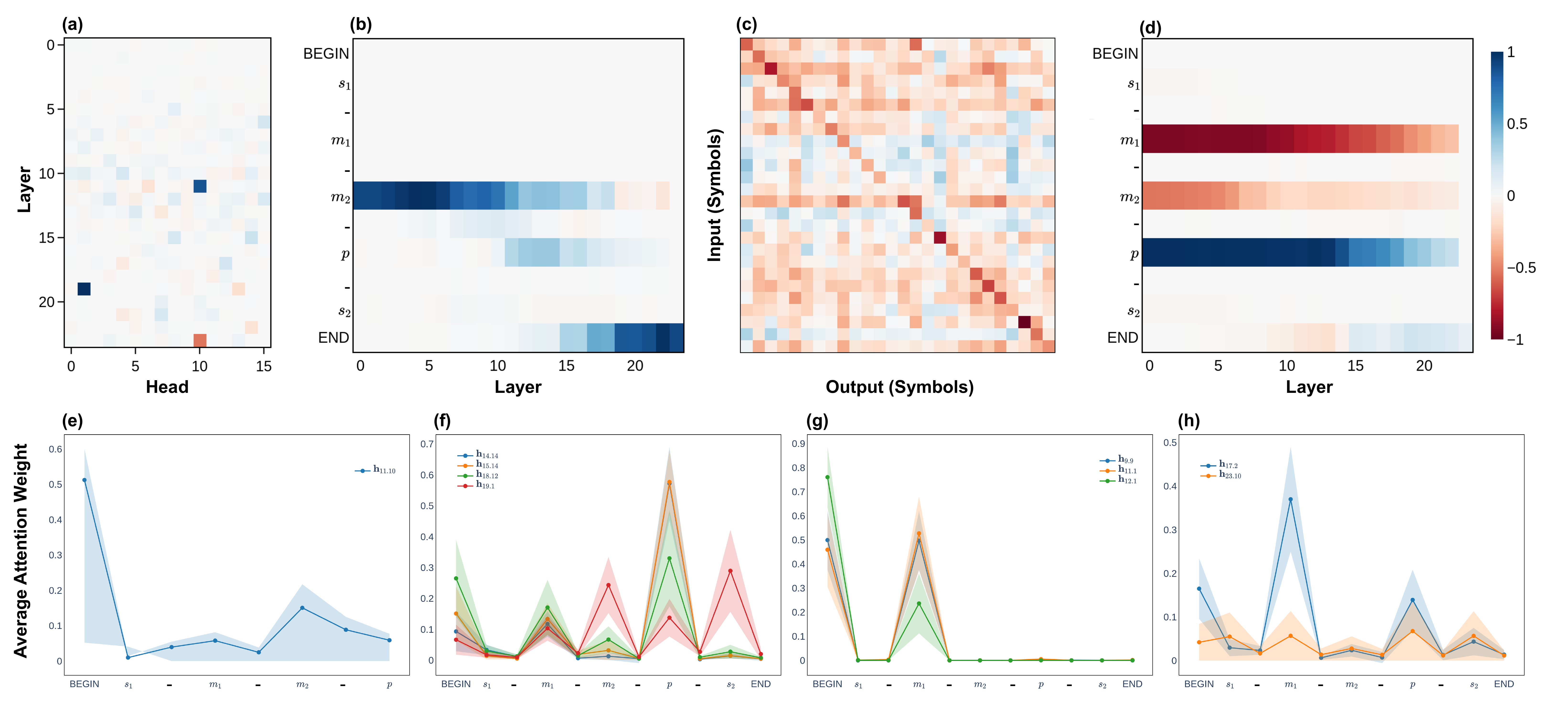}
\caption{
Visualization of the symbolic reasoning circuit identified in the model.  
(a) Attention output patching and (b) residual stream patching using middle-term interventions highlight the role of attention head $\mathbf{h}_{11.10}$ in aggregating information from the duplicated middle term to the predicate token position $[p]$.  
(c) Logit lens visualization of the OV (output vector) of $\mathbf{h}_{11.10}$ across 26 uppercase letters reveals its suppressive effect on the attended token at $[p]$.  
(d) Residual stream patching with all-term corruption indicates that key reasoning information is concentrated at term token positions.  
(e)–(h) Average attention weights across the batch:  
(e) attention from token position $[p]$, attending strongly to $[m_2]$;  
(f)–(h) attention from the last token position, identifying which heads (mover heads) propagate term information to the output.  
For clarity, a dash (–) on the axis indicates the averaged values for tokens appearing between terms.
}
\label{fig:symbolic_circuit_analysis}
\end{figure*}

\subsection{Causal Interventions}
The choice of tokens to corrupt is a critical aspect of the experimental design, and it is essential to establish an appropriate type of intervention that aligns with the hypothesis being tested. In this study, we employ two distinct interventions to isolate the mechanisms related to the reasoning schema from the propagation of specific token information:

\paragraph{(1) Middle-Term Corruption.} To investigate the reasoning mechanism employed in the syllogism completion task, we primarily focus on the transitive property of the middle term, $(m_1 \rightarrow m_2) \wedge (m_1 \equiv m_2)$. We hypothesize that disrupting the transitive property will localize the component responsible for syllogistic reasoning. Therefore, our intervention method replaces the second middle term $m_2$ with an unseen symbol $m_2'$, breaking the equality and effectively corrupting the validity of the reasoning, pivoting the correct answer towards $m$: $(s, m_1, m_2, p) \rightarrow (s, m_1, m_2', p)$.
    
\paragraph{(2) All-Term Corruption.} We examine how term-related information flows to the last position for the final prediction to identify potential mover heads. To this end, we replace the original terms $(s, m_1, m_2, p)$ with different terms $(s', m_1', m_2', p')$ while keeping the answers unchanged. We hypothesize that if heads carry the information preferring $p$ over $m$ for the valid prediction, the logit difference will increase; otherwise, it will decrease.

\subsection{Circuit Ablation and Evaluation}
Once we discover a relevant circuit $\mathcal{C}$, we evaluate its necessity and sufficiency using mean ablation. Specifically, we define a circuit $\mathcal{C}$ as necessary if ablating the identified heads $\mathcal{H}$ in $\mathcal{C}$ and preserving the remaining components in $\mathcal{M}$ decreases the original performance. Conversely, we define a circuit $\mathcal{C}$ as sufficient if preserving only the identified heads $\mathcal{H}$ in $\mathcal{C}$ and ablating all the remaining heads in the original model $\mathcal{M}$ is sufficient to obtain the performance of $\mathcal{M}$. To perform mean ablation, we gradually average the target attention heads from downstream to upstream layers and measure the average logit difference $\delta(p, m)$ to assess the impact of the ablation on $\mathcal{M}$. Additional details are included in Appendix~\ref{app:appendix_D}.

\section{Mechanistic Interpretation}
\label{sec:empirical_evaluation}

\paragraph{Empirical Setup.} We select the LM for circuit analysis based on a trade-off between model size and performance. To this end, we measure the accuracy of GPT-2 \citep{Radford2019} at various sizes (117M, 345M, 762M, and 1.5B) on the syllogism completion task, aiming to identify potential phase transition points where performance changes significantly~\citep{Jason2022, Kaplan2020}. We observe a marked transition from small to medium sizes, where average performance across three datasets increases by 439.06\% (see Figure~\ref{fig:model_comparison}(a), Appendix~\ref{app:appendix_E}). The shift is even more pronounced under the logit difference metric (see Section~\ref{sec:generalisation_models} and Figure~\ref{fig:model_comparison}(b), Appendix~\ref{app:appendix_E}). Based on these findings, we select \textit{GPT-2 Medium}—whose architecture is summarized in Appendix~\ref{app:appendix_F}—for circuit discovery, and subsequently evaluate generalization across model sizes and architectures (Sections~\ref{sec:generalisation_size} and~\ref{sec:generalisation_models}).

\subsection{Transitive Reasoning Mechanisms} 

\paragraph{Middle-term corruption reveals information flow relevant to the transitive property.}
Figure~\ref{fig:symbolic_circuit_analysis}(a-b) presents the results of the middle-term intervention targeting attention heads and residual stream. Figure~\ref{fig:symbolic_circuit_analysis}(a), in particular, reveals the positive role of heads $\mathbf{h}_{11.10}$ and $\mathbf{h}_{19.1}$ (where $\mathbf{h}_{l.k}$ denotes the $k$th head in layer $l$). Moreover, we observe an information propagation pattern from the $[m_2]$ position to the $[p]$ position (Figure~\ref{fig:symbolic_circuit_analysis}(b)). These observations allow us to hypothesize that information from $[m_2]$ is conveyed to $[p]$ on the residual stream by attention head $\mathbf{h}_{11.10}$, which exhibits a strong patching score around the layer responsible for the propagation. To verify this hypothesis, we further investigate the attention weights between $[p]$ and $[m_2]$. As expected, $[m_2]$ is the position most highly attended by $[p]$, with an average attention weights of $0.15 \pm 0.07$ (Figure~\ref{fig:symbolic_circuit_analysis}(e)). These results suggest that information is indeed moved from $[m_2]$ to $[p]$ on the residual stream subspace, with $\mathbf{h}_{11.10}$ playing a crucial role in this mechanism.

\paragraph{Duplicate middle-term information is aggregated via induction heads.}
We further investigate how information from $[m_2]$ is moved to $[p]$, positing that this relates to the model's internal reasoning mechanism. We notice that at position $[p]$, the model can observe the complete AAA-1 syllogistic structure (Figure~\ref{fig:methodology}(a)) with middle-term duplication. We hypothesize that this structural information for reasoning is collected at one position, given that information refinement occurs at the specific position such as last token~\citep{Hanna2024, stolfo2023}. To verify this, we employ path patching \citep{Wang2023}, a more selective variant of activation patching, to trace the information flow of head $\mathbf{h}_{11.10}$. Our results show that $\mathbf{h}_{11.10}$ operates based on several induction heads \citep{Elhage2021} ($\mathbf{h}_{5.8}$, $\mathbf{h}_{6.1}$, $\mathbf{h}_{6.15}$ and $\mathbf{h}_{7.2}$) formed at $[m_2]$. These heads attend to the $[[m_1] +1]$ token due to the $m_1 \equiv m_2$ conditioned matching operation, likely containing $m_1$-related information from previous token heads. We conclude, therefore, that $m_1$ and $m_2$ information are aggregated at position $[p]$. Additional details of the path patching results are available in the Appendix~\ref{app:appendix_G}.

\paragraph{A suppressive mechanism is revealed through Logit Lens.}
To better understand the internal mechanism occurring at the $[p]$ position, we investigate attention head $\mathbf{h}_{11.10}$. Having previously examined the attention pattern (composed of attention weights), we now focus on the attention value and output by analyzing the OV circuit ($W_V W_O$)~\citep{Elhage2021} using the logit lens method~\citep{Nostalgebraist2020}. Interestingly, the result in Figure~\ref{fig:symbolic_circuit_analysis}(c) shows a clear negative diagonal pattern suggesting that $\mathbf{h}_{11.10}$ strongly suppresses the logit when attending to the same token as the corresponding output. 
Given our previous findings that $\mathbf{h}_{11.10}$ reads information from the subspace of token $[m_2]$, we conclude that it applies a suppressive mechanism to $m$-related information and writes it back to the residual stream's subspace at $[p]$. If later heads carry this information to the last token, this mechanism becomes crucial for the model to arrive at the correct answer. For simplicity, we name head $\mathbf{h}_{11.10}$ as $m$-suppression head. 

\subsection{Term-Related Information Flow} 

\paragraph{Key information is moved from term-specific positions to the last position.}
Now, we use the all-term intervention method to localize mover heads that carry term-related information. In the residual stream patching results (Figure~\ref{fig:symbolic_circuit_analysis}(d)), we observe the highest positive score at the $[p]$ position, while negative scores are most prominent at the $[m_1]$ and $[m_2]$ positions, indicating that the information residing in these positions indeed contributes to their corresponding token prediction. This observation aligns with the importance of token embedding information at their respective positions given iterative refinement on the residual stream~\citep{Simoulin2021}. A closer examination of the last token position reveals that the negative effect propagates from relatively early layers (approximately layer 10 onwards), yet positive effects are from later layers (approximately layer 14 onwards), incurring a positive shift of the model's prediction from $m$ to $p$.

\paragraph{Information is carried by later positive and negative mover heads.}
Given the findings that the information for prediction resides in term-specific positions, we trace which attention heads transfer the information from each term position ($[p]$, $[m_1]$ and $[m_2]$) to the last token. We call these heads mover heads as the existence of ``positive or negative mover heads'' that carry or suppress information from the specific token position to the last token position~\citep{Wang2023, Garcia2024}. To identify the sources of information flow, we apply attention value-based patching and isolate nine notable mover heads (see Appendix~\ref{app:appendix_H} for localization details). Among them, $\mathbf{h}_{14.14}$, $\mathbf{h}_{15.14}$, and $\mathbf{h}_{18.12}$ act as positive copy heads, attending strongly to $[p]$, while $\mathbf{h}_{19.1}$ functions as a positive suppression head, exhibiting high attention to $[m_2]$ and $[s]$ (see Figure~\ref{fig:symbolic_circuit_analysis}(f)). In contrast, $\mathbf{h}_{9.9}$, $\mathbf{h}_{11.1}$, $\mathbf{h}_{12.1}$, $\mathbf{h}_{17.2}$, and $\mathbf{h}_{23.10}$ serve as negative copy heads, focusing primarily on $[m_1]$ (Figure~\ref{fig:symbolic_circuit_analysis}(g)) or showing more diffuse pattern (Figure~\ref{fig:symbolic_circuit_analysis}(h)).


\paragraph{A long induction mechanism in early layer movers causes biases towards the middle term.}
Notably, the early negative mover heads ($\mathbf{h}_{9.9}$, $\mathbf{h}_{11.11}$ and $\mathbf{h}_{12.1}$ in Figure~\ref{fig:symbolic_circuit_analysis}(g)) exhibit significantly high attention to $[m_1]$. This pattern indicates that these heads transfer information from the $[m_1]$ position directly to the last token position, negatively affecting the final prediction and biasing the model towards $m$. We notice that this behavior closely resembles that of induction heads~\citep{Elhage2021} defined on bi-grams ($[s]$$[\text{are}]$) rather than uni-gram ($[\text{are}]$), supported by previous token heads in upstream layers (e.g., $\mathbf{h}_{8.1}$). 
\subsection{Circuit Discovery Summary}
Overall, we found that the mechanisms for syllogistic inference involve the following phases:

\paragraph{(1) Long Induction: } Early layers exhibit biases towards the wrong conclusion due to long-range repetition of the first premise ``All A are B''.

\paragraph{(2) Duplication:} Induction heads aggregate information about duplicated middle terms in the premises.

\paragraph{(3) Suppression:} The model aggregates and inhibits middle-term information (i.e.,`B') suppressing the long induction mechanism.

\paragraph{(4) Mover:} Token-specific information is propagated to the last position. The process ends in the prediction shift from `B’ to the correct token ‘C’.

These results suggest that the circuit is characterized by an internal error correction mechanism. Interestingly, this mechanism is different from the way human experts would reason on syllogistic arguments through the systematic application of abstract logical primitives and inference rules.

\section{Circuit Evaluation} 
\begin{figure}[t]
\centering
\includegraphics[width=\linewidth]{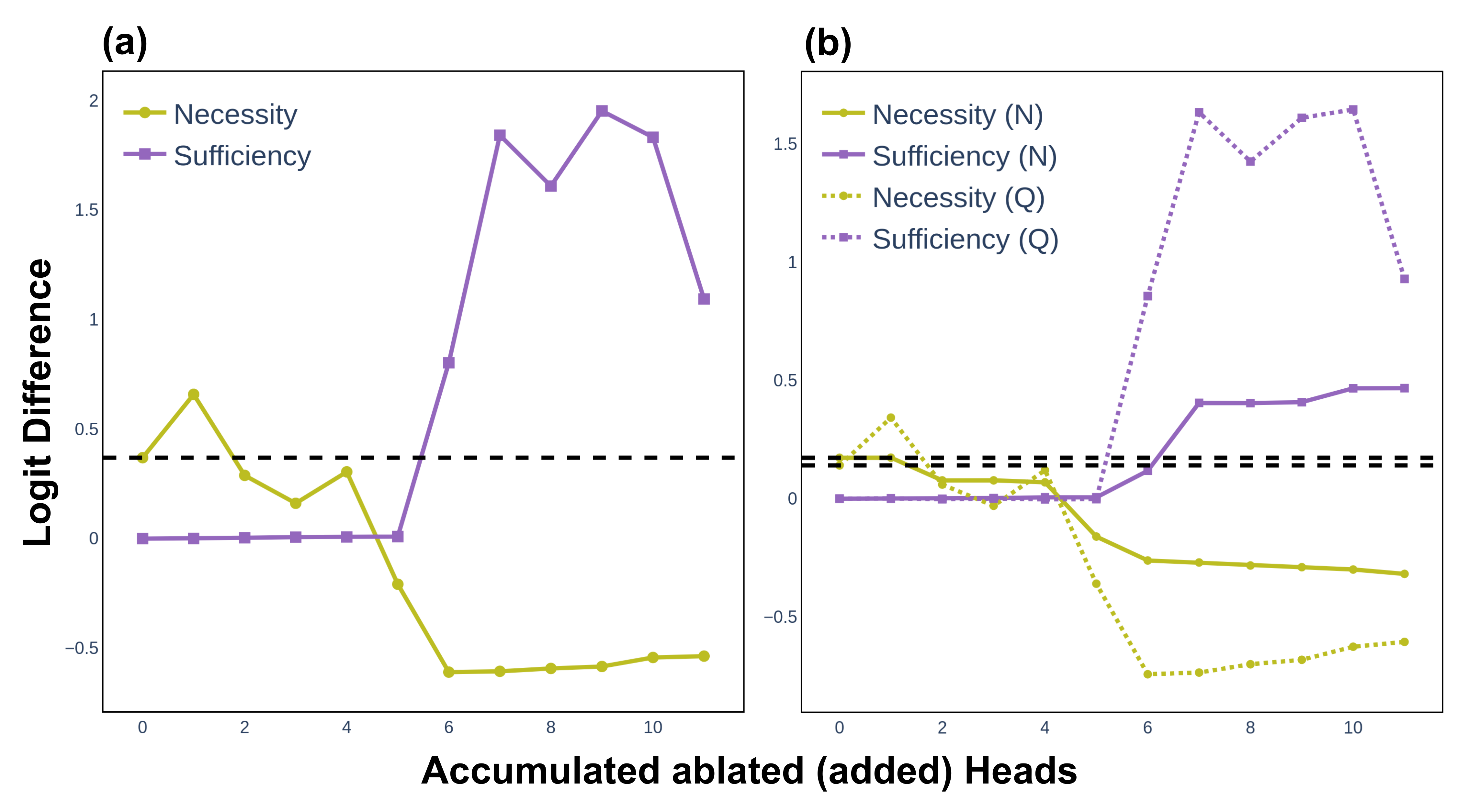}
\caption{Circuit ablation results for (a) correctness and (b) robustness on the symbolic dataset. Solid lines show numeric perturbation-based performance, dotted lines indicate quantifier perturbation-based performance in (b), and the dashed line shows the baseline logit difference without knockouts. These results validate the necessity, sufficiency, and robustness of the identified symbolic reasoning circuit.}
\label{fig:evaluation}
\end{figure}

\paragraph{The circuit is sufficient and necessary for symbolic arguments.}To evaluate the comprehensive correctness of the circuit, we assess necessity and sufficiency via the ablation method described in section~\ref{fig:methodology}. The ablation study in Figure~\ref{fig:evaluation}(a) shows that the identified circuit is both necessary and sufficient, demonstrating a consistent performance degradation when removing circuit components and revealing a complete restoration of the original model's performance when considering only the circuit's subcomponents. 

\begin{figure}[t!]
\centering
\includegraphics[width=\linewidth]{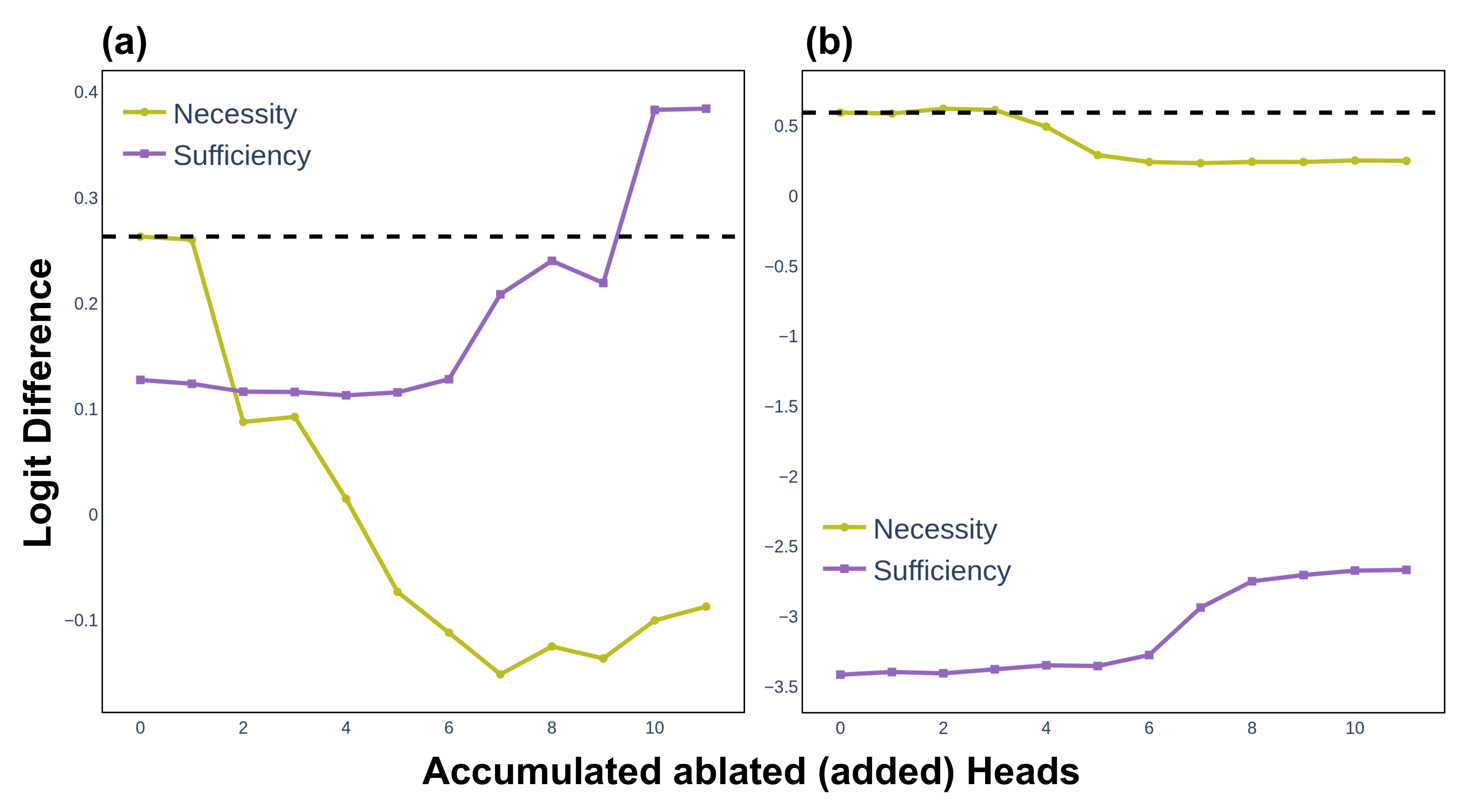}
\caption{
Circuit ablation results for evaluating transferability on non-symbolic datasets: (a) the belief-consistent and (b) the belief-inconsistent. The dashed line represents the baseline logit difference without knockouts. In both cases, the symbolic circuit is necessary, as ablating it reduces performance. However, the circuit is only sufficient for belief-consistent data, as recovery fails in belief-inconsistent settings, suggesting interference from belief-bias.}
\label{fig:transferability}
\end{figure}

\paragraph{The circuit is robust to superficial variations.} We further verify the robustness of the symbolic circuit to superficial and semantic-preserving perturbations. In particular, we modify the letters into numbers (Figure~\ref{fig:evaluation}(b)) and adopt semantically equivalent quantifiers and related verbs (e.g., ``All ... are'' is converted into ``Each ... is''). The ablation result demonstrates that both types of perturbations do not undermine the sufficiency and necessity property of the circuit.

\subsection{Circuit Transferability to Concrete Arguments}

\paragraph{The circuit is necessary for non-symbolic arguments, yet not sufficient for belief-inconsistent ones.}
We present ablation results for the two generated non-symbolic datasets: belief-consistent (Figure \ref{fig:transferability}(a)) and belief-inconsistent (Figure \ref{fig:transferability}(b)). Notably, the symbolic circuit proves necessary for both types of non-symbolic inputs, suggesting that the logic derived from symbolic syllogisms remains essential even when natural words are substituted. Regarding sufficiency, while belief-consistent data show significant performance recovery, we observe an inability to restore performance on the belief-inconsistent set despite an increasing trend. These results indicate that belief biases encoded in different attention heads may play an important role.

\paragraph{Belief biases corrupt reasoning mechanisms.}
To further investigate the impact of belief biases, we again conduct intervention experiments. Specifically, we observe that in the AAA-1 syllogism schema $(s, m_1, m_2, p)$, the subject token $s$ should be irrelevant for deriving the correct answer $p$ over $m$. 
Therefore, we leverage this property to verify whether word-specific biases are introduced when instantiating the schema with concrete knowledge. To this end, we perform activation patching by corrupting the subject term, transforming $(s, m_1, m_2, p)$ to $(s', m_1, m_2, p)$, and measuring the degradation in logit difference as $\delta_{+}(p, m) - \delta_{-}(p, m)$. Notably, the non-symbolic setting exhibits a significant performance degradation of $0.66 \pm 1.27$, representing a 299.96\% change from the baseline. In contrast, the symbolic setting shows a minimal degradation of $0.00 \pm 0.64$, a mere 0.35\% drop. This drastic difference supports our hypothesis that the knowledge acquired during pre-training corrupts the content-independent reasoning circuit identified on the symbolic set with additional attention heads.

\subsection{Generalization to Syllogistic Schemes}

\begin{table}[t!]
\small
\centering
\begin{tabular}{lcccc}
\toprule
\textbf{Mood-Figure} & \textbf{C1} & \textbf{C2} & \textbf{C3} & \textbf{Acc} \\
\midrule
\textbf{AII-3 (\textit{Datisi})} & $\checkmark$ & $\checkmark$ & $\checkmark$ & \textbf{0.84} \\
\textbf{IAI-3 (\textit{Disamis})} & $\checkmark$ & $\checkmark$ & $\checkmark$ & \textbf{0.68} \\
\textbf{IAI-4 (\textit{Dimaris})} & $\checkmark$ & $\checkmark$ & $\checkmark$ & \textbf{0.68} \\
\textbf{AAA-1 (\textit{Barbara})} & $\checkmark$ & $\checkmark$ & $\checkmark$ & \textbf{0.67} \\ \midrule
\textbf{EAE-1 (\textit{Celarent})} & $\checkmark$ & - & 
$\checkmark$ & \textbf{0.59} \\ 
\textbf{EIO-4 (\textit{Fresison})} & $\checkmark$ & - & $\checkmark$ & \textbf{0.53} \\
\textbf{EIO-3 (\textit{Ferison})} & $\checkmark$ & - & $\checkmark$ & \textbf{0.53} \\ \midrule
AII-1 (\textit{Darii}) & $\checkmark$ & $\checkmark$ & - & 0.43 \\
AOO-2 (\textit{Baroco}) & - & - & - & 0.24 \\
AEE-4 (\textit{Camenes}) & - & - & - & 0.24 \\
OAO-3 (\textit{Bocardo}) & $\checkmark$ & - & - & 0.22 \\
EIO-1 (\textit{Ferio}) & $\checkmark$ & - & - & 0.18 \\
EIO-2 (\textit{Festino}) & - & - & - & 0.09 \\
EAE-2 (\textit{Cesare}) & - & - & - & 0.08 \\
AEE-2 (\textit{Camestres}) & - & - & - & 0.04 \\
\bottomrule
\end{tabular}
\caption{Generalizability across unconditionally valid syllogistic forms. Columns C1, C2, and C3 indicate whether conditions are met ($\checkmark$) or not (-): C1: Necessity, C2: Sufficiency, C3: Positive Logit Difference. `Accuracy (Acc)' shows performance on the completion task for each syllogism. The results show that the model achieves high downstream accuracy particularly for affirmative syllogisms.}
\label{tab:syllogism_variation}
\end{table}
We further aim to understand whether the symbolic circuit is specific to the AAA-1 syllogism (\textit{Barbara}). To this end, we extend our experiment to encompass all 15 unconditionally valid syllogisms (see Appendix~\ref{app:appendix_A} and~\ref{app:appendix_I} for details about the syllogistic schemes and more experimental results). To evaluate the transferability to all 15 schemes, we verify three main conditions, reported in Table~\ref{tab:syllogism_variation}: (C1) Necessity, (C2) Sufficiency, and (C3) Positive Logit Difference. We also measure the accuracy of the completion task for each syllogism. From Table~\ref{tab:syllogism_variation}, we observe that the circuit is sufficient and necessary for all the syllogistic schemes in which the model achieves high downstream accuracy ($\geq$ 60\%), supporting the conclusion that the circuit includes components that are crucial for the emergence of syllogistic reasoning in general. Notably, all three conditions are satisfied for the affirmative syllogisms (AII-3, IAI-3, IAI-4), for which the model achieves an accuracy above 0.6. 

\subsection{Generalization to Model Sizes}
\label{sec:generalisation_size}
Next, we expand our analysis to different sizes of GPT-2 (small, large and XL).
Here, the intervention results show similar suppression mechanism patterns and information flow across all models. However, the residual stream of GPT-2 small in the all-term corruption setup is reversed due to its low downstream accuracy (see Appendix~\ref{app:appendix_J} for additional details). Moreover, as the model size increases, we found evidence that the contribution of attention heads becomes more complex. We hypothesize this might be caused by a stronger impact of world knowledge with increasing size, as suggested by the decrease in accuracy on the symbolic dataset and the increase on the non-symbolic set observed for the XL model (see Figure \ref{fig:model_comparison}(a)--(b), Appendix~\ref{app:appendix_E}).

\subsection{Generalization to Model Families}
\label{sec:generalisation_models}
We further extend our investigation to assess whether a similar reasoning mechanism can be identified across different LMs. Specifically, we evaluate the circuit on: Pythia with 70M, 160M, 410M, and 1B parameters~\citep{pythia}; Qwen 2.5 with 0.5B and 1.5B parameters~\citep{qwen2.5}; Llama 3.2 with 1B parameters~\citep{llama3}; and additionally, GPT-2-medium after being fine-tuned on argumentative texts for critical thinking~\citep{Betz2021}.

Our results reveal distinct grouping patterns. In relatively small models—such as Pythia 70M and 160M—we observe erratic activation patterns that mirror the limitations seen in under-scaled models like GPT-2 small. In contrast, models with more than 0.5B parameters (namely, Pythia 410M, Pythia 1B, and Llama 3.2 1B) exhibit a compatible pattern in terms of information flow and suppressive mechanisms. Qwen exhibits a variant suppressive mechanism, with the suppression head operating at the last token position rather than $[p]$. Meanwhile, the fine-tuned GPT-2 medium model yields activation patterns that are nearly indistinguishable from its pre-trained counterpart, suggesting that the core reasoning circuit originates from pre-training rather than fine-tuning. Table~\ref{tab:model_comparison_accuracies} summarizes the accuracy of each model across the Symbolic, Belief-Consistent, and Belief-Inconsistent datasets (see further circuit analysis details in Appendix~\ref{app:appendix_K}).

Overall, our results indicate that while the reasoning circuit generalizes effectively across sufficiently scaled models, subtle architectural nuances can lead to distinct but comparable dynamics.

\begin{table}[t!]
    \centering
    \small
    \begin{tabular}{l|c c c}
        \toprule
        \textbf{Model} & \textbf{S} & \textbf{BC} & \textbf{BI} \\
        \midrule
        \multicolumn{4}{l}{\textbf{Group 1: Erratic Pattern}} \\
        \midrule
        Pythia-70M & 0.19 & 0.02 & 0.02 \\
        Pythia-160M & 0.00 & 0.12 & 0.21 \\
        \midrule
        \multicolumn{4}{l}{\textbf{Group 2: Compatible Pattern}} \\
        \midrule
        Pythia-410M & 0.99 & 0.40 & 0.51 \\
        Pythia-1B & 0.30 & 0.31 & 0.43 \\
        Llama-3.2-1B & 1.00 & 0.81 & 0.70 \\
        \midrule
        \multicolumn{4}{l}{\textbf{Group 3: Variant Pattern}} \\
        \midrule
        Qwen2.5-0.5B & 1.00 & 0.96 & 0.90 \\
        Qwen2.5-1.5B & 1.00 & 0.99 & 0.90 \\
        \bottomrule
    \end{tabular}
    \caption{
    Accuracy comparison across different models: 
    \textbf{symbolic (S)}, 
    \textbf{belief-consistent (BC)}, and 
    \textbf{inconsistent (BI)}. 
    \textbf{Group 1} models exhibit erratic patterns, failing to establish a stable reasoning mechanism. 
    \textbf{Group 2} models exhibit a compatible reasoning circuit with consistent information flow and suppression. 
    \textbf{Group 3} exhibits a functional but distinct suppression mechanism, where suppression is applied at the last token position.}
    \label{tab:model_comparison_accuracies}
\end{table}

\section{Related Work}
Mechanistic circuit analysis has emerged as a promising approach to interpreting the internal mechanisms of Transformers \citep{Olah2020, Nanda2023, Olsson2022,Wang2023, Garcia2024, Hanna2024}.
Existing approaches investigating internal reasoning mechanisms mainly focus on math-related tasks, elucidating the information flow for answering mathematical questions \citep{stolfo2023} and examining arithmetic operations \citep{Quirke2024}. 
Most pertinent to our work, \citet{Wiegreffe2024} provides a mechanistic interpretation of multiple-choice question answering, investigating attention head-level patterns. 
In general, our mechanistic analysis complements recent work investigating the challenges in processing reasoning arguments that contradict established beliefs and whether the reasoning in LMs stems from internalized rules or memorized content \citep{Ando2023,Yu2023, Wu2024,talmor2020,Wu2024,Kassner2020,Haviv2023, Feldman2020,Monea2024, Yu2023,Singh2020,eisape-etal-2024-systematic}.
To address this question, different mechanistic approaches have been adopted to localizing factual associations~\citep{Meng2022, Geva2023, Dai2022} or assessing conditions for generalization~\citep{Wang2024}. However, to the best of our knowledge, we are the first to investigate mechanisms for syllogistic reasoning.

\section{Conclusion}
In this study, we presented a comprehensive mechanistic interpretation of syllogistic reasoning in language models. By combining activation patching, embedding space analysis, and circuit ablation, we uncovered a structured error-correction mechanism characterized by the suppression of duplicated middle terms and the propagation of relevant information via mover heads. Our findings show that this circuit is necessary and sufficient for symbolic syllogistic reasoning. Moreover, we demonstrated that, while this reasoning mechanism generalizes across syllogistic schemes, model architectures and sizes, it remains susceptible to contamination from belief biases when instantiated with natural language inputs. Overall, our findings suggest that LMs learn transferable reasoning mechanisms but that, at the same time, such mechanisms might be contaminated and suppressed by the same world knowledge acquired during pre-training. Further studies will be required to understand how the discovered symbolic circuit can be effectively disentangled from world knowledge to enable systematic generalization in LMs' reasoning. 

\section{Limitations}

It is important to acknowledge some of the limitations of our study. First, our analysis is conducted predominantly on the transitive property and term-specific information and does not consider the full spectrum of the reasoning dynamics that might appear in more complex scenarios which are hard to model via causal intervention techniques. Second, Our experimentation is confined to specific syllogistic reasoning templates due to the intricate level of granularity and the computational complexity required in circuit analysis. Despite these limitations, however, we believe our findings could offer valuable insights into the reasoning mechanisms adopted by language models for formal reasoning and related biases, laying a solid foundation for future research in this field. 

\section*{Acknowledgements}
This work was funded by the Swiss National Science Foundation (SNSF) project ``NeuMath'' (200021\_204617),  by the CRUK National Biomarker Centre, and supported by the Manchester Experimental Cancer Medicine Centre and the NIHR Manchester Biomedical Research Centre.

\bibliography{custom}

\appendix
\setcounter{section}{0}

\section{Unconditionally Valid Syllogism Schemes}
We list all 15 unconditionally valid syllogisms (Table~\ref{tab:uncon_valid_syllogisms}).
\label{app:appendix_A}

\begin{table*}[h!]
\centering
\begin{tabular}{cccp{3cm}p{3cm}p{3cm}}
\toprule
\textbf{Name} & \textbf{Mood} & \textbf{Figure} & \textbf{Premise (m)} & \textbf{Premise (M)} & \textbf{Conclusion} \\ 
\midrule
Barbara  & AAA & 1 & $\forall x (s, m)$ & $\forall x (m, p)$ & $\forall x (s, p)$ \\
       &      &    & \textit{All A are B.} & \textit{All B are C.} & \textit{All A are C.} \\ 
Celarent & EAE & 1 & $\forall x (s, m)$ & $\forall x \neg(m, p)$ & $\forall x \neg(s, p)$ \\
         &     &   & \textit{All A are B.} & \textit{No B are C.} & \textit{No A are C.} \\ 
Darii    & AII & 1 & $\exists x (s, m)$ & $\forall x (m, p)$ & $\exists x (s, p)$ \\
         &     &   & \textit{Some A are B.} & \textit{All B are C.} & \textit{Some A are C.} \\ 
Ferio    & EIO & 1 & $\exists x (s, m)$ & $\forall x \neg(m, p)$ & $\exists x \neg(s, p)$ \\
         &     &   & \textit{Some A are B.} & \textit{No B are C.} & \textit{Some A are not C.} \\ 
\midrule
Camestres & AEE & 2 & $\forall x \neg(s, m)$ & $\forall x (p, m)$ & $\forall x \neg(s, p)$ \\
          &     &   & \textit{No A are B.} & \textit{All C are B.} & \textit{No A are C.} \\ 

Cesare    & EAE & 2 & $\forall x (s, m)$ & $\forall x \neg(p, m)$ & $\forall x \neg(s, p)$ \\
          &     &   & \textit{All A are B.} & \textit{No C are B.} & \textit{No A are C.} \\ 

Baroco    & AOO & 2 & $\exists x \neg(s, m)$ & $\forall x (p, m)$ & $\exists x \neg(s, p)$ \\
          &     &   & \textit{Some A are not B.} & \textit{All C are B.} & \textit{Some A are not C.} \\ 

Festino   & EIO & 2 & $\exists x (s, m)$ & $\forall x \neg(p, m)$ & $\exists x \neg(s, p)$ \\
          &     &   & \textit{Some A are B.} & \textit{No C are B.} & \textit{Some A are not C.} \\ 
\midrule
Disamis   & IAI & 3 & $\forall x (m, s)$ & $\exists x (m, p)$ & $\exists x (s, p)$ \\
          &     &   & \textit{All B are A.} & \textit{Some B are C.} & \textit{Some A are C.} \\ 

Datisi    & AII & 3 & $\exists x (m, s)$ & $\forall x (m, p)$ & $\exists x (s, p)$ \\
          &     &   & \textit{Some B are A.} & \textit{All B are C.} & \textit{Some A are C.} \\ 

Ferison   & EIO & 3 & $\exists x (m, s)$ & $\forall x \neg(m, p)$ & $\exists x \neg(s, p)$ \\
          &     &   & \textit{Some B are A.} & \textit{No B are C.} & \textit{Some A are not C.} \\ 

Bokardo   & OAO & 3 & $\forall x (m, s)$ & $\exists x \neg(m, p)$ & $\exists x \neg(s, p)$ \\
          &     &   & \textit{All B are A.} & \textit{Some B are not C.} & \textit{Some A are not C.} \\ 
\midrule
Dimaris   & IAI & 4 & $\forall x (m, s)$ & $\exists x (p, m)$ & $\exists x (s, p)$ \\
          &     &   & \textit{All B are A.} & \textit{Some C are B.} & \textit{Some A are C.} \\ 

Camenes   & AEE & 4 & $\forall x \neg(m, s)$ & $\forall x (p, m)$ & $\forall x \neg(s, p)$ \\
          &     &   & \textit{No B are A.} & \textit{All C are B.} & \textit{No A are C.} \\ 

Fresison  & EIO & 4 & $\exists x (m, s)$ & $\forall x \neg(p, m)$ & $\exists x \neg(s, p)$ \\
          &     &   & \textit{Some B are A.} & \textit{No C are B.} & \textit{Some A are not C.} \\ 
\bottomrule
\end{tabular}
\caption{15 Unconditionally valid syllogism schemes. The table lists the syllogisms by their traditional names, moods, and figures, with formalized logical expressions on the first line and corresponding natural language examples on the second line. The minor premise (m) is presented before the major premise (M) to emphasize the transitive property in AAA-1 syllogism.}
\label{tab:uncon_valid_syllogisms}
\end{table*}

\section{Dataset Generation}
\label{app:appendix_B}

\subsection{Symbolic}
The symbolic dataset comprises sentences where all terms in the premises are represented by abstract symbols (uppercase alphabet letters). From the set of all 26 uppercase alphabets, three-letter triples (e.g., A, B and C) are randomly sampled. For each triple, six permutated prompts and label pairs are generated following syllogism templates designed to minimise latent semantic interference among alphabet symbol tokens.

\subsection{Non-symbolic}
The non-symbolic datasets are constructed based on GenericsKB \citep{Bhakthavatsalam2020}, a resource that provides a foundation for the evaluation of sentence veracity with associated truthfulness scores (0-1). The dataset construction process involves the following steps:

\begin{itemize}
\item \textbf{Extraction:} Select generic sentences with a truthfulness score of 1 based on GenericsKB~\citep{Bhakthavatsalam2020}, specifically those in the form \textit{A are B}, using regular expression.
    
\item \textbf{Syllogism Construction:} We form universal affirmative syllogism arguments (\textit{Barbara}) based on a template by chaining sentences where the predicate of one sentence logically matches the subject of another.

\item \textbf{Constraints:} We exclude terms tokenized into multiple tokens to maintain consistency in comparison with the symbolic dataset, which is essential for coherent circuit analysis.

\item \textbf{Classification:} To address issues of partial inclusion and syntactic ambiguity in constructed syllogism arguments, we employ GPT-4 
\citep{Achiam2023} to classify whether arguments contain only truthful premises and whether the middle-terms are syntactically and semantically equivalent. This step helps us classify a belief-consistent dataset and a belief-inconsistent dataset.

\item \textbf{Validation:} We manually evaluate the alignment of classified arguments with human belief-consistency (i.e., premises are true and logic is valid).
    
\end{itemize}
This process ensures that our non-symbolic dataset maintains logical equivalence with the symbolic dataset while incorporating meaningful semantic real-word concepts. 

\subsection{Data Statistics}

For all experiments, we use 90 samples each for both symbolic and non-symbolic arguments to balance analytical depth with computational efficiency. This sample size sufficiently yields statistically significant activation patching results ($p < 0.05$). We organize data statistics for generated datasets (Table~\ref{tab:data_statistics}).
\begin{table}[h!]
\centering

\begin{tabular}{lccc}
\toprule
\textbf{Statistic} & \textbf{SYM} & \textbf{BC} & \textbf{BI} \\
\midrule
\textbf{Number of Samples} & 90 & 90 & 90 \\
\textbf{Token Sequence Length} & 15 & 15 & 15 \\
\textbf{Unique Terms ($s$)} & 26 & 87 & 86 \\
\textbf{Unique Terms ($m$)} & 26 & 70 & 35 \\
\textbf{Unique Terms ($p$)} & 26 & 59 & 64 \\
\bottomrule
\end{tabular}
\caption{Summary of Dataset Statistics. \textbf{SYM} refers to the symbolic dataset, \textbf{BC} refers to the non-symbolic belief-consistent dataset, and \textbf{BI} refers to the non-symbolic belief-inconsistent dataset.}
\label{tab:data_statistics}
\end{table}

\section{Embedding Space Analysis Details}
\label{app:appendix_C}
One established method for analyzing hidden activation in transformer language models is the logit lens, which projects activation into embedding space~\citep{Nostalgebraist2020, Elhage2021, Geva2022}. We focus on an input-agnostic approach~\citep{Dar2023,Hanna2024}, utilizing both unembedding ($W_U$) and embedding ($W_E$) matrices to construct a $\mathbb{R}^{|V| \times |V|}$ matrix, where $|V|$ denotes the model's vocabulary size. This study employs the OV circuit \citep{Elhage2021} of attention heads, formed by attention value and output weights ($W_VW_O$), to understand how source tokens generally influence output logits. The OV circuit-based logit lens for attention head $\mathbf{h}$ is formalized as $W_EW^{\mathbf{h}}_VW_O^{\mathbf{h}}W_U \in \mathbb{R}^{|V| \times |V|}$. Following the previous works~\citep{Elhage2021, Dar2023}, this formulation omits layer normalization.

\section{Circuit Ablation Method Details}
\label{app:appendix_D}
To measure the necessity and sufficiency of the circuit $\mathcal{C}$ in the model $\mathcal{M}$, we knock out attention heads in $\mathcal{H}$ from the model $\mathcal{M}$ and measure the average logit difference $\delta(p, m)$ along the batch. We denote the logit difference in circuit state $\mathcal{C}$ as $\delta(p, m, \mathcal{C})$ and every subset of heads set as $H \subset \mathcal{H}$. 

In order to verify the head's necessity in the model, we conduct a cumulative ablation of $\mathcal{C}$ from total circuit $\mathcal{M}$, progressing from downstream to upstream layers. At each ablation step \(k\), where $\mathcal{C}_k^* = \mathcal{M} \setminus H_k$ and $H_k$ denotes the set of ablated attention heads up to step $k$ as: 
\begin{align*}
\mathbb{E}_{x\sim X}[\delta_k(p, m, \mathcal{C}_k^*)] \quad \text{where} \quad \mathcal{C}_k^* = \mathcal{M} \setminus H_k
\end{align*}

Conversely, we perform a cumulative addition of attention heads for evaluating the sufficiency of the circuit, starting from earlier layers and progressing to later ones, while maintaining all other attention heads in a mean-ablated state. At each addition step $j$, where $\mathcal{C}_j^* = \mathcal{M} \setminus (\mathcal{H} \setminus H_j)$ and $H_j$ denotes the set of added attention heads up to step $j$ as:
\begin{align*}
\mathbb{E}_{x\sim X}[\delta_j(p, m, \mathcal{C}_j^*)] \quad \text{where} \quad \mathcal{C}_j^* = \mathcal{M} \setminus (\mathcal{H} \setminus H_j)
\end{align*}

\section{Phase Transition}
\label{app:appendix_E}
\begin{figure*}[ht!]
\centering
\includegraphics[width=1.0\linewidth]{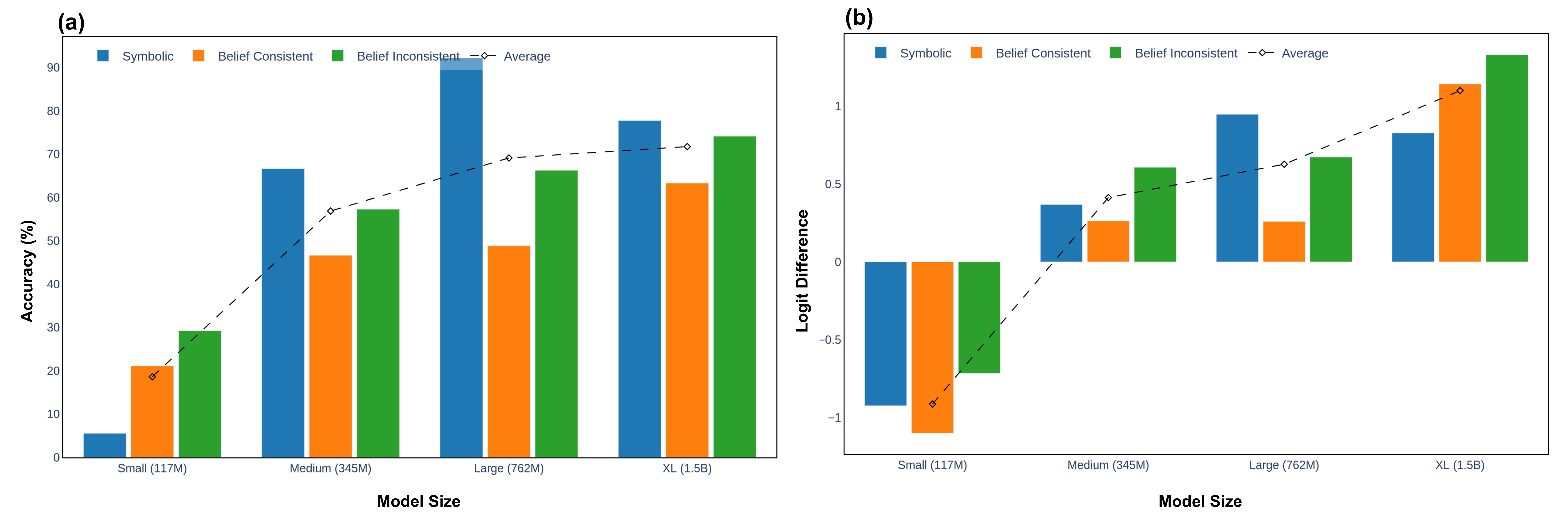}
\caption{
Accuracy and logit difference on the syllogism completion task across all different sizes of GPT-2 for three different datasets.}
\label{fig:model_comparison}
\end{figure*}

In Figure~\ref{fig:model_comparison}(a--b), we observe a notable phase transition from small to
medium models in accuracy and logit difference.

\section{GPT-2 Model Architecture}
\label{app:appendix_F}
We provide a self-contained concise overview of the GPT-2 model architecture, highlighting the main components and their mathematical relationships. Bias terms are not presented for the simplicity. We refer the conventions of notation in~\citet{Elhage2021} and~\citet{Geva2023}.

\paragraph{Notation}
\begin{itemize}
    \item \( x_i \) - One-hot encoded vector representing the \( i \)-th token in the input sequence.
    \item \( p_i \) - One-hot encoded vector representing the \( i \)-th positional information.
    \item \( r_i^l \) - Residual stream vector representing the \( i \)-th token at layer $l$ in the input sequence.
    \item \( W_E^{\text{token}}, W_E^{\text{pos}} \) - Token and positional embedding matrices.
    \item \( A_i^l \) - Output from the Multi-Head Self-Attention sublayer
    \item \( M_i^l \) - Output from Multi-Layer Perceptron (MLP) sublayer.
    \item \( W_Q, W_K, W_V, W_O \) - Query, key, value, output weight matrices of attention heads.
    \item \( W^{\text{in}}, W^{\text{out}} \) - Input and output weight matrices of MLP feed-forward networks.
    \item \( d \) - Dimension of the attention head state embedding vectors.
    \item \( D \) - Dimension of the hidden state embedding vectors.
    \item \( N \) - Length of the input sequence.
    \item \( L \) - Total number of transformer layers.
    \item \( \mathcal{V} \) - Vocabulary set of the model.
    \item \( \sigma \) and \( \sigma' \) - Activation functions used in self-attention and MLP sublayers, respectively.
\end{itemize}

\paragraph{Embedding Layer.}
Each token in the input sequence is transformed into an embedded representation by combining token-specific and positional embeddings:
\begin{equation*}
r_i^0 = x_i W_E^{\text{token}} + p_i W_E^{\text{pos}}
\end{equation*}
This operation initializes the input for the transformer layers, where \( r_i^0 \) represents the initial embedded state of the \( i \)-th token.

\paragraph{Residual Stream.}
The residual stream facilitates the propagation of information across transformer layers. It is updated at each layer by contributions from the multi-head self-attention and multi-layer perceptron sublayers:
\begin{equation*}
r_i^l = r_i^{l-1} + A_i^l + M_i^l
\end{equation*}

\paragraph{Multi-Head Self-Attention Sublayer.}
This sublayer processes information by applying self-attention mechanisms across multiple `heads' of attention, enabling the model to capture various aspects of the input data, where $\sigma$ represent the non-linearity function:
\begin{equation*}
A_i^l = \sum_{h \in H} \left(\sigma\left(\frac{(r_i^{l} W^h_Q)(r_i^{l} W^h_K)^T}{\sqrt{d_K}}\right)(r_i^{l} W^h_V)\right)W^h_O
\end{equation*}
Each head computes a weighted sum of all tokens' transformed states, focusing on different subsets of sequence information.

\paragraph{Multi-Layer Perceptron Sublayer.}
Each token's representation is further processed in a position-wise manner by the MLP sublayer:
\begin{equation*}
M_i^l = \sigma'(r_i^l W^{in})(W^{out})^T
\end{equation*}
The MLP modifies each token's state locally, enhancing its ability to process information.

\paragraph{Layer Normalization.}
Before processing by self-attention and MLP processing, each token's state is normalized to stabilize learning and reduce training time:
\begin{equation*}
\text{LN}(r_i^l) = \gamma \odot \frac{r_i^l - \mu_i^l}{\sqrt{(\sigma_i^l)^2 + \epsilon}} + \beta
\end{equation*}
This step ensures that the activations across different network layers maintain a consistent scale.

\paragraph{Prediction Head.}
The prediction head generates logits for the next token prediction using the final transformed states:
\begin{equation*}
\text{logits}_N = \text{LN}(r_N^L) W_U
\end{equation*}
The predicted token is chosen based on the sampling method (e.g., greedy decoding) at the last position $N$. 

\section{Path Patching Details}
\label{app:appendix_G}
\begin{figure}[h!]
    \centering
    \includegraphics[width=0.8\linewidth]{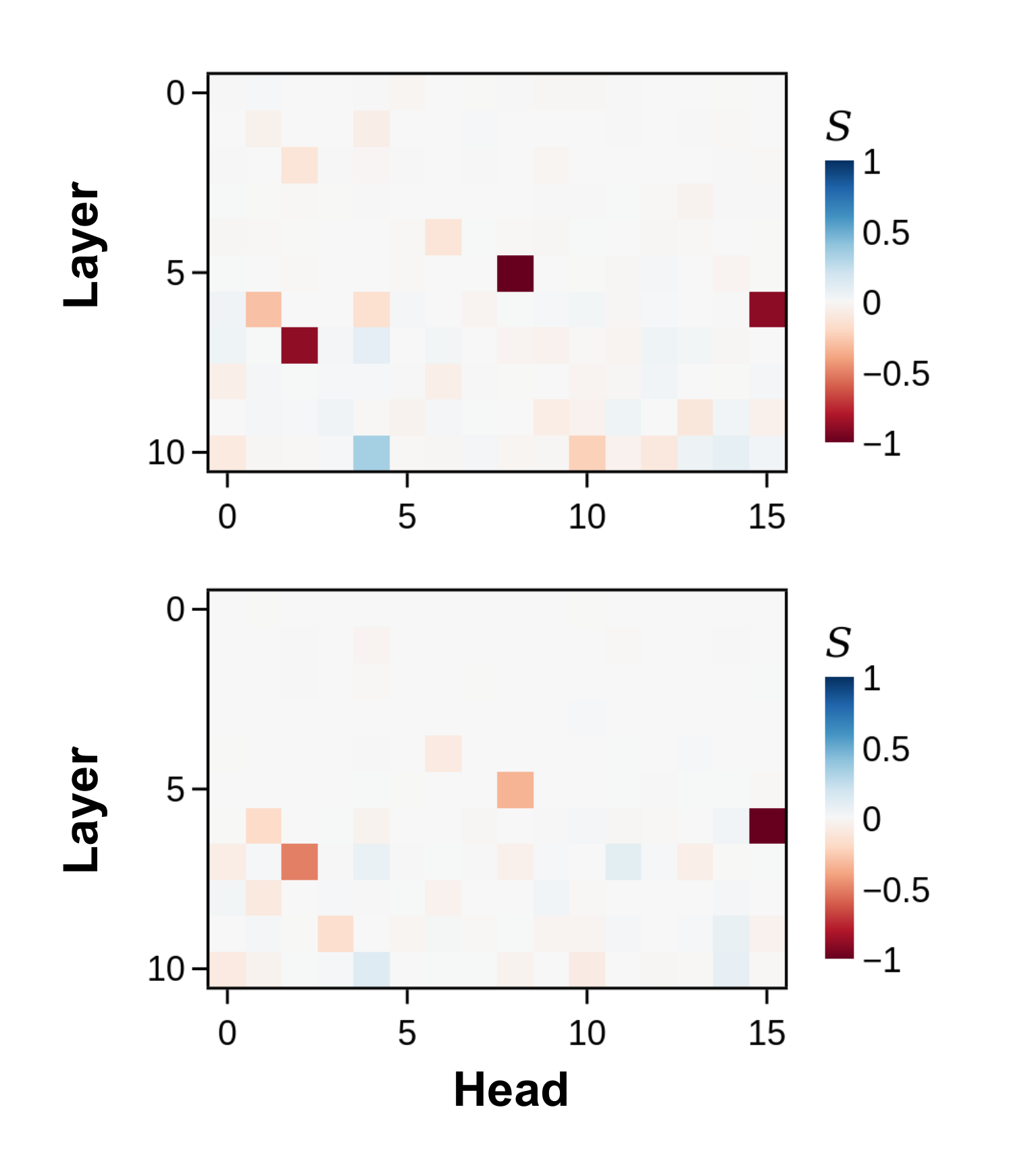}
    \caption{Path patching results for (Top) symbolic inputs and non-symbolic inputs (Bottom) in the middle-term corruption setup. Note that negative heads are positively influencing heads, as we replace corrupted activation with clean ones in our noising method.}
    \label{fig:path_patching_result}
\end{figure}

Path patching, an generalized version of activation patching, aims to compute direct effects among model components rather than indirect effects diffused from intervened components \citep{Wang2023}. This method involves freezing non-targeted activations during an initial forward pass, storing the targeted activation state in intervention, and then executing a subsequent forward pass with the targeted activation state substituted by the stored one. This approach isolates targeted component-to-component effects, eliminating non-relevant interactions.
In our implementation, we employ a noising method where clean activations are replaced with corrupted ones, resulting in negative scores indicating positive contributions from corresponding components. Our findings, illustrated in Figure \ref{fig:path_patching_result}, reveal that $\mathbf{h}_{11.10}$ strongly operates based on several heads: $\mathbf{h}_{5.8}$, $\mathbf{h}_{6.1}$, $\mathbf{h}_{6.15}$, and $\mathbf{h}_{7.2}$. Manual investigation of attention patterns subsequently confirmed these as induction heads \citep{Elhage2021}.

\section{Localizing Mover Heads}
\label{app:appendix_H}

To efficiently classify the numerous attention heads (384 in GPT-2 medium) within the model, we calculate the Positional Patching Difference (PPD) to construct a four-quadrant analysis:
\begin{equation*}
    \text{PPD} = |S[p]| - |S[m_1] + S[m_2]|
\end{equation*}

This method involves intervening in the attention value weights $W_V$ and constructs a distribution layout of heads, providing a systematic classification for quadrant groups. The layout is composed of the PPD score on the x-axis and the patching score for all sequence positions ($\mathcal{S}$) on the y-axis.

Based on this PPD score quadrant layout analysis, we categorize heads according to their quadrant positions, revealing insights into their functional roles:
\begin{itemize}
\item[(1)] First quadrant (Positive Copy Candidates): Heads with $\mathcal{S} > 0$ and positive PPD, predominantly copying $[p]$-based information.
\item[(2)] Second quadrant (Positive Suppression Candidates): Heads with $\mathcal{S} > 0$ but negative PPD, predominantly suppressing $[m]$-based information.
\item[(3)] Third quadrant (Negative Copy Candidates): Heads with $\mathcal{S} < 0$ and negative PPD, predominantly copying $[m]$-based information.
\item[(4)] Fourth quadrant (Negative Suppression Candidates): Heads with $\mathcal{S} < 0$ but positive PPD, predominantly suppressing $[p]$-based information.
\end{itemize}

The intervention of attention head values is selected because all-term corruption maintains the same positional alignment for all samples $(s, m_1, m_2, p) \rightarrow (s', m_1', m_2', p')$, resulting in minimal impact from attention head pattern interventions composed by attention query and key. Consequently, we can infer that the attention head output patching results primarily are derived from the attention value activation. This property enables us to localize the source token positions from which the term information is moved by attention value patching.

Formally, when we denote the attention query, key, value, and output weight matrices as $W_Q, W_K, W_V$, and $W_O$ respectively, and the residual stream vector as $r$, one attention head output can be expressed as: $\sigma\left( (rW_Q)( rW_K)^T \right)(rW_V)W_O$, where $\sigma$ represents the non-linearity function (softmax in GPT-2) applied in the attention pattern.

To include only influential heads in the circuit, we extracted outlier heads with an absolute patching score exceeding the threshold ($\tau$), defined by the mean and standard deviation of all heads' scores as:
\begin{equation*}
\{ \mathbf{h} \mid |\mathcal{S}_{\mathbf{h}}| > \tau \} \quad \text{where} \quad \tau = \mu + 2\sigma
\end{equation*}

\begin{figure*}[h!]
    \centering
    \includegraphics[width=0.85\linewidth]{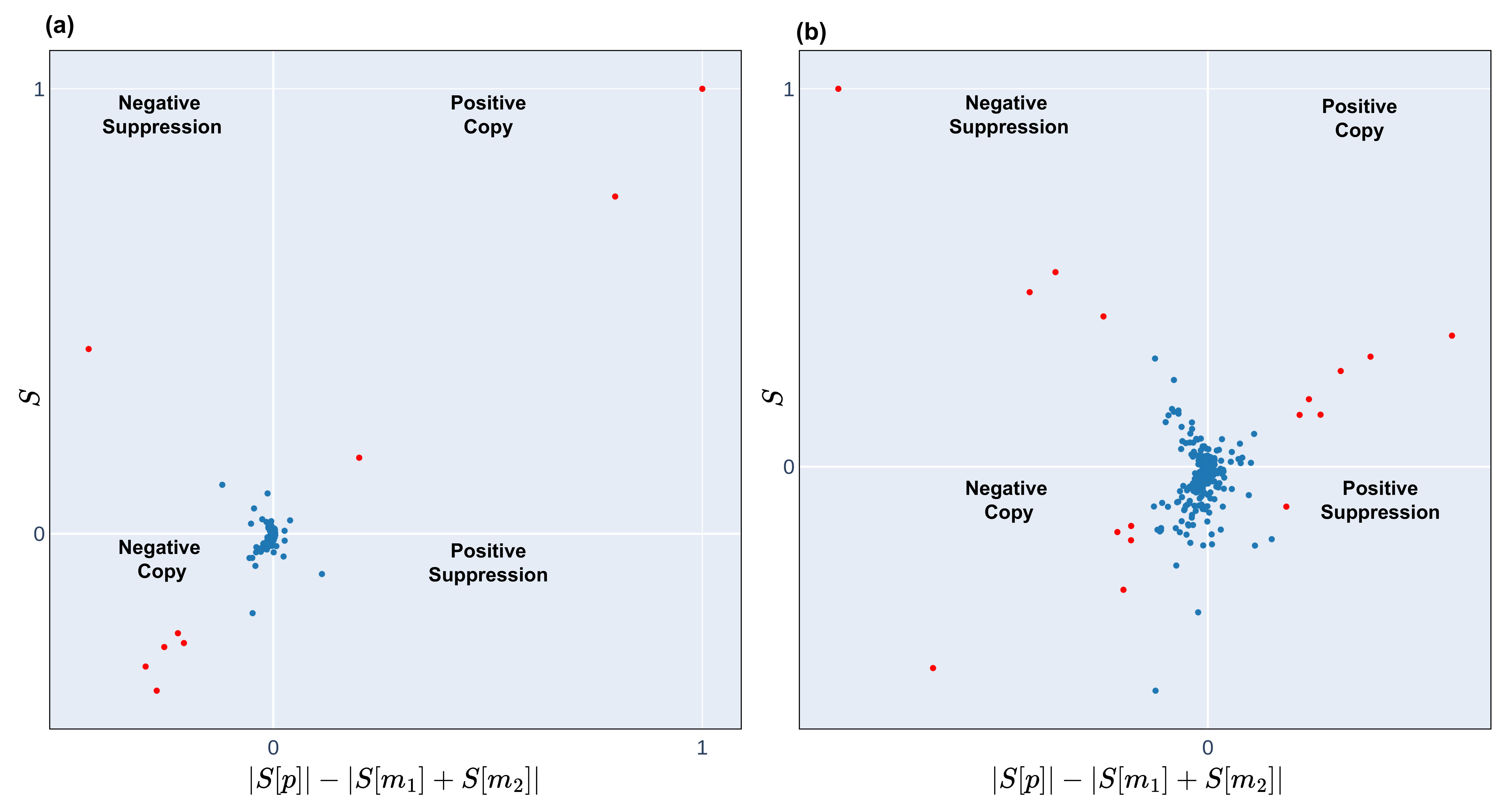}
    \caption{Distribution of attention heads in PPD-based quadrant analysis. (a) Results for symbolic inputs. (b) Results for non-symbolic inputs. The x-axis represents normalized PPD scores, while the y-axis represents normalized attention value-based patching scores ($S$). Red dots indicate attention heads above the threshold ($\tau$) of the patching score.}
    \label{fig:mover_localisation}
\end{figure*}

Our analysis identifies 9 heads for the symbolic circuit, as illustrated by red dots in Figure \ref{fig:mover_localisation}(a).  $\mathbf{h}_{14.14}$, $\mathbf{h}_{15.14}$, and $\mathbf{h}_{18.12}$ are classified as positive copy candidates, while $\mathbf{h}_{19.1}$ is grouped as a positive suppression candidates. $\mathbf{h}_{9.9}$, $\mathbf{h}_{11.1}$, $\mathbf{h}_{12.1}$, $\mathbf{h}_{17.2}$, and $\mathbf{h}_{23.10}$ are classified as negative copy candidates. It is noteworthy that non-symbolic results reveal a more complex and noisy pattern in the mover localization process (Figure \ref{fig:mover_localisation}(b)).

\section{Generalizability to Other Syllogisms}
\label{app:appendix_I}
\begin{figure*}[h!]
\centering
\includegraphics[width=0.9\linewidth]{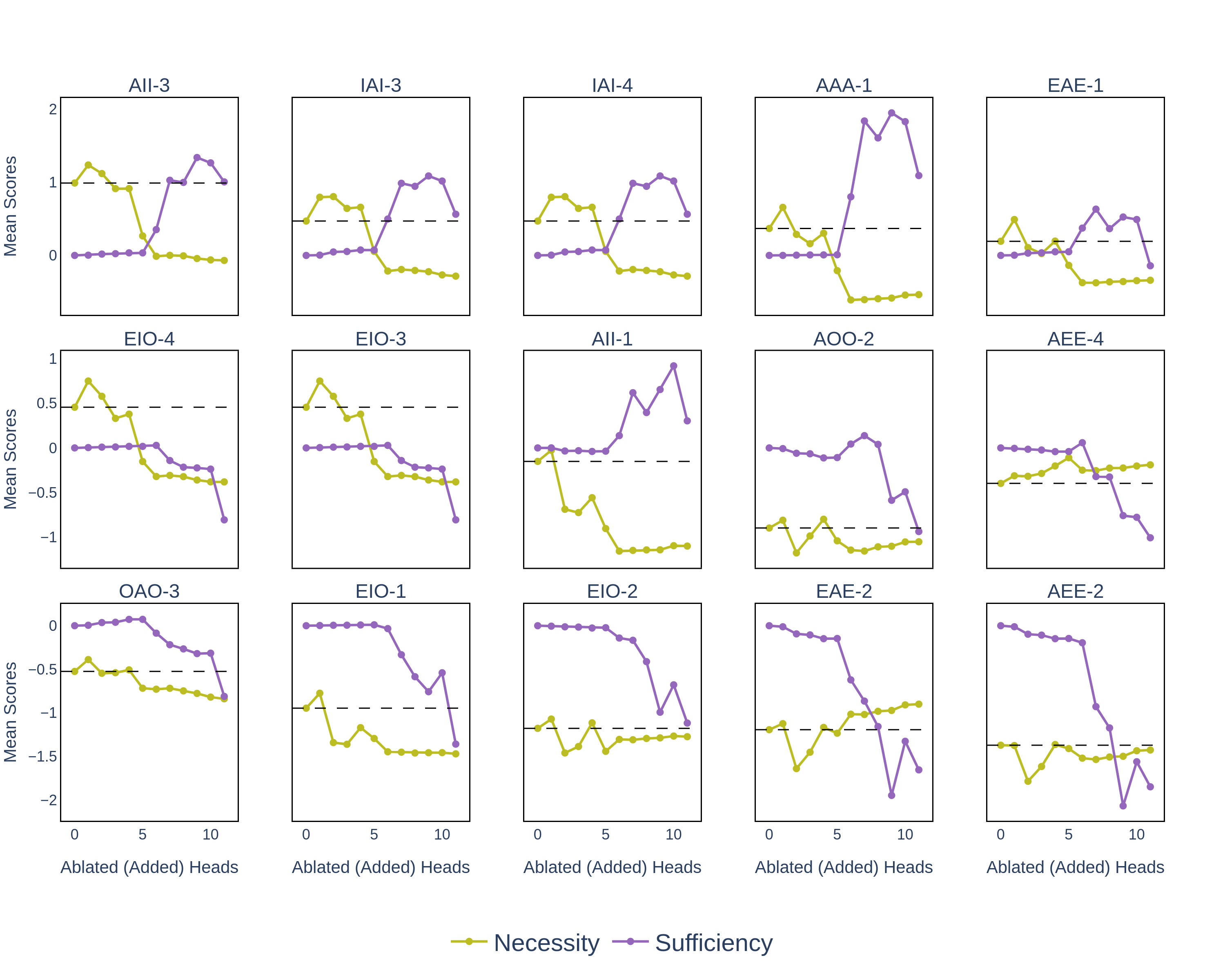}
\caption{Necessity and sufficiency performance results from circuit ablation method for all unconditionally valid syllogistic forms. Labels (e.g., AII-3) denote mood and figure combinations.}
\label{fig:syllogism_generalisability}
\end{figure*}

To assess the generalizability of our model, we applied the circuit ablation method to all 15 unconditionally valid syllogisms (Table~\ref{tab:uncon_valid_syllogisms}), and the result is shown in Figure~\ref{fig:syllogism_generalisability}. 
Unconditionally valid syllogisms are logical arguments that maintain their validity irrespective of the truth values of their premises. The validity of these syllogisms is determined solely by their logical form, which is characterized by mood and figure combinations. The mood of a syllogism is defined by the arrangement of four proposition types ("All (A)", "No (E)", "Some (I)", "Some ... not (O)") across its two premises and conclusion. Conversely, the figure of a categorical syllogism is determined by the structural arrangement of terms within the constituent sentences. This approach enables the evaluation of syntactic logic in isolation from contextual interference and belief-bias effects.

\section{Generalizability Across Model Sizes}
\label{app:appendix_J}
\begin{figure*}[h!]
\centering
\includegraphics[width=1.0\linewidth]{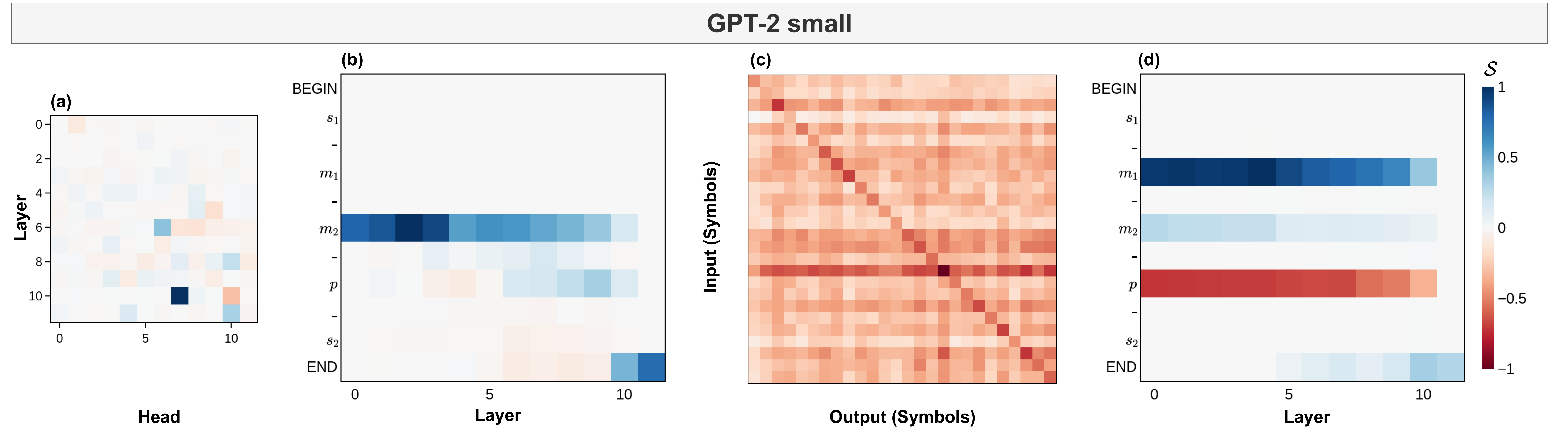}
\par\bigskip
\includegraphics[width=1.0\linewidth]{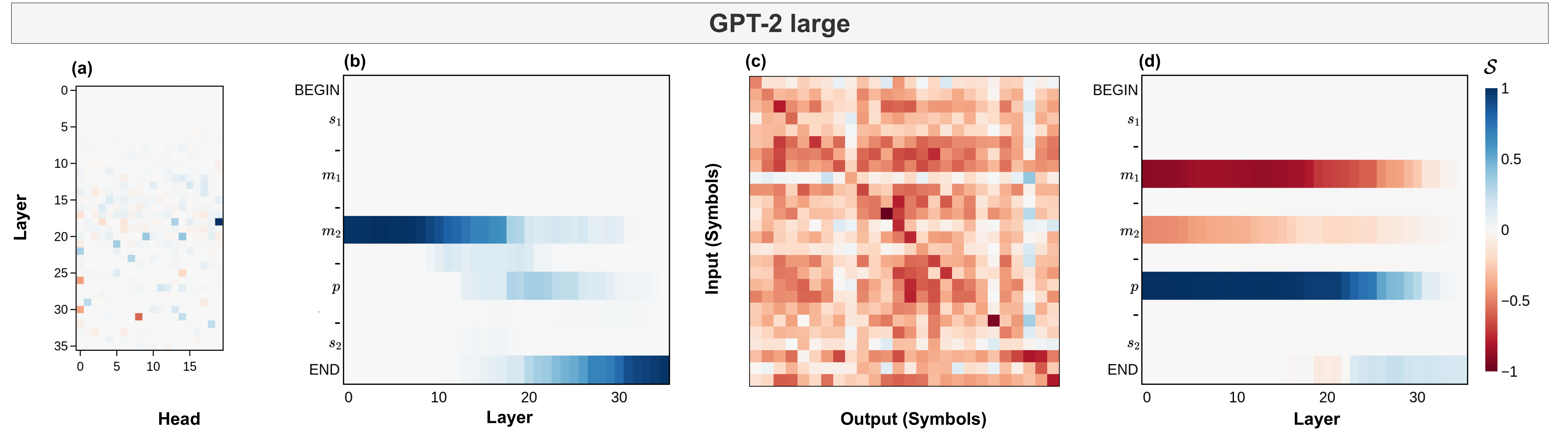}
\par\bigskip
\includegraphics[width=1.0\linewidth]{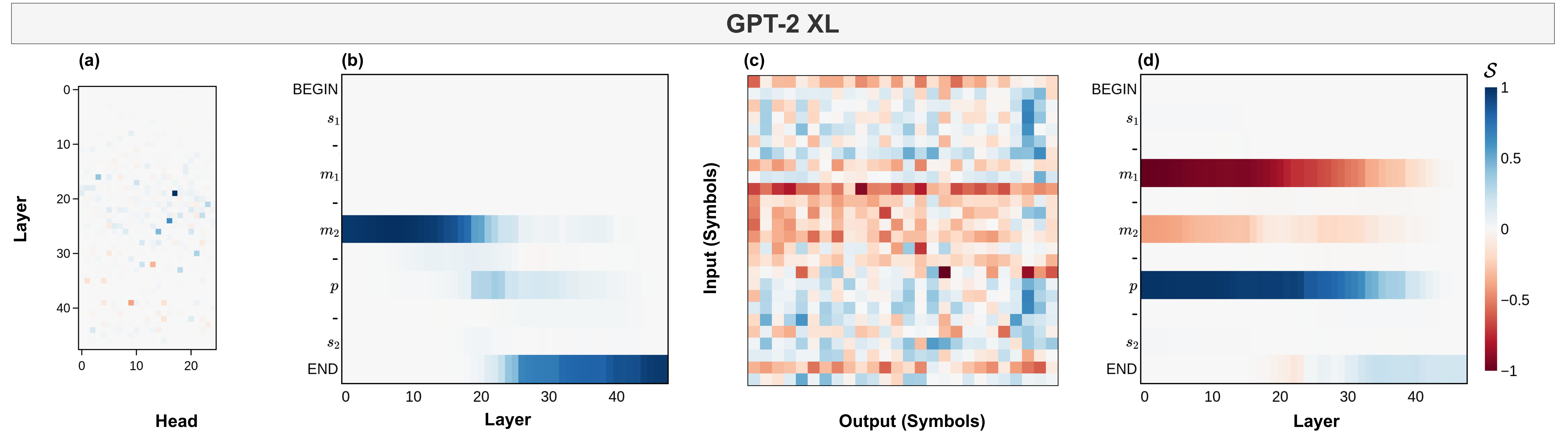}
\caption{Comprehensive results of the symbolic circuit analysis across different model sizes. (a) Attention output patching results and (b) residual stream patching results in the middle-term intervention setup. (c) OV circuit logit lens results for $m$-suppression head, with input and output comprising 26 uppercase letters. (d) Residual stream patching results in the all-term corruption setup. For clarity, a dash (–) indicates the averaged values for tokens appearing between terms.}
\label{fig:model_generalisability}
\end{figure*}

Our approach is guided by the universality hypothesis, which suggests that similar representational patterns and circuits emerge across models trained on the same dataset~\citep{Olah2020, Kornblith2019, li2015}. We assume that core mechanisms should persist across these different model sizes. Figure \ref{fig:model_generalisability} presents the results of this comparative analysis.

\section{Generalizability Across Different Models}
\label{app:appendix_K}
\begin{figure*}[h!]
\centering
\includegraphics[width=1.0\linewidth]{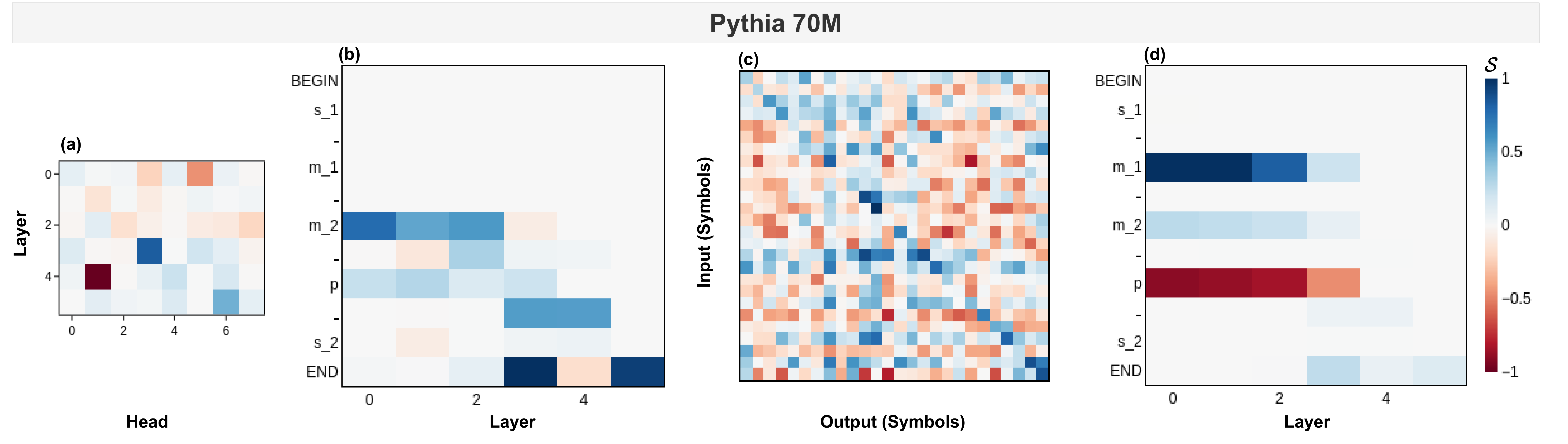}
\par\bigskip
\includegraphics[width=1.0\linewidth]{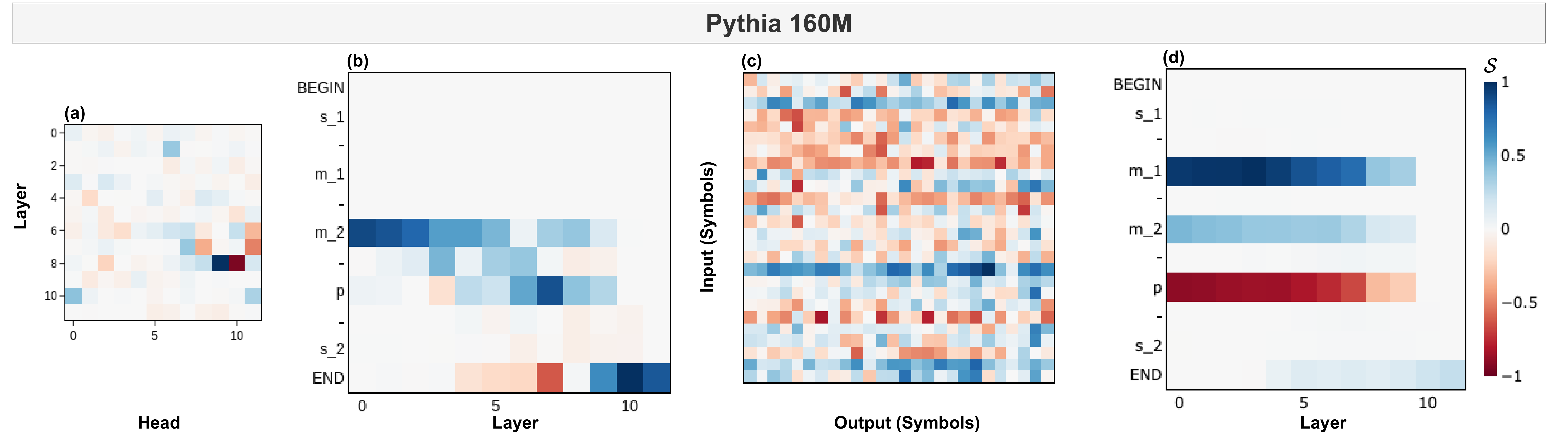}
\par\bigskip
\includegraphics[width=1.0\linewidth]{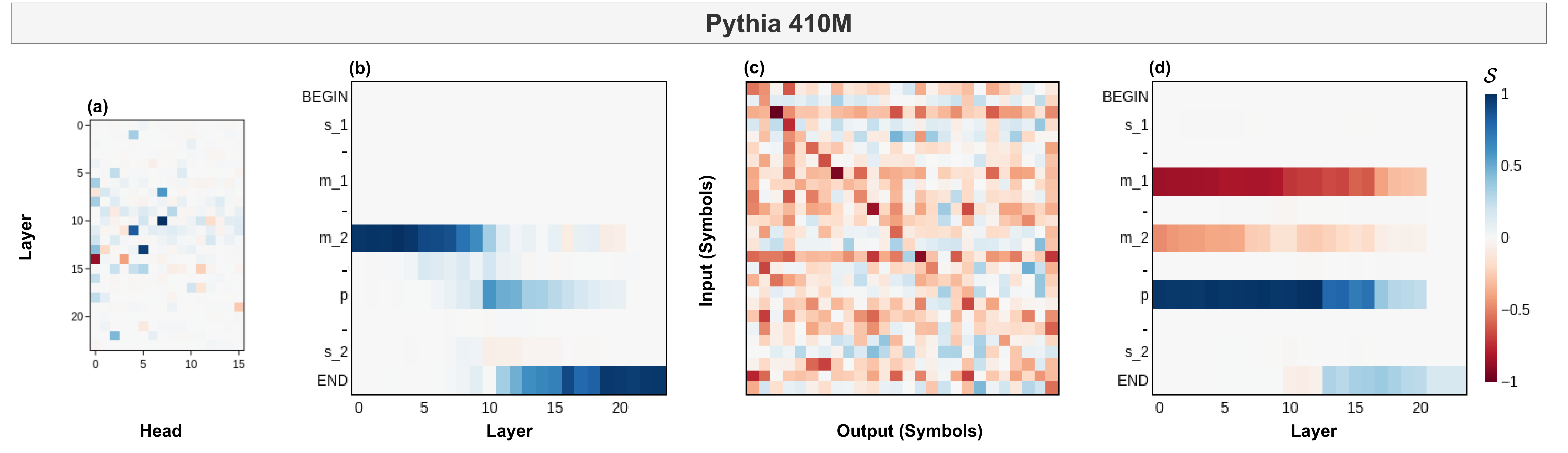}
\par\bigskip
\includegraphics[width=1.0\linewidth]{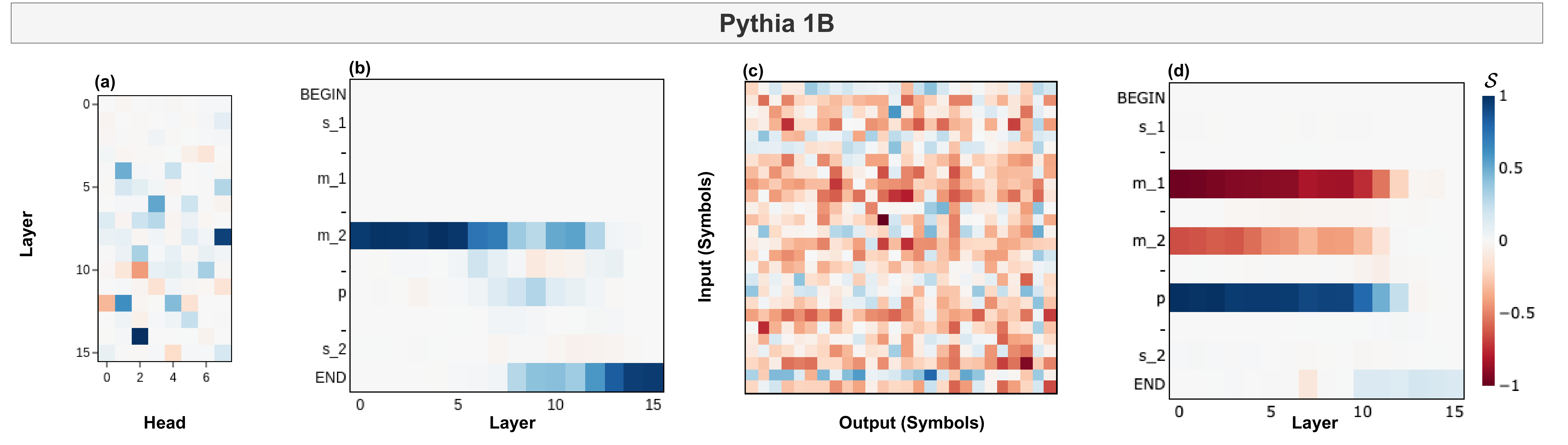}
\caption{Comprehensive results of the symbolic circuit analysis across different series of models (Pythia). (a) Attention output patching results and (b) residual stream patching results in the middle-term intervention setup. (c) OV circuit logit lens results for $m$-suppression head, with input and output comprising 26 uppercase letters. (d) Residual stream patching results in the all-term corruption setup. For clarity, a dash (–) indicates the averaged values for tokens appearing between terms.}
\label{fig:different_model_generalisability}
\end{figure*}

\begin{figure*}[h!]
\centering
\includegraphics[width=1.0\linewidth]{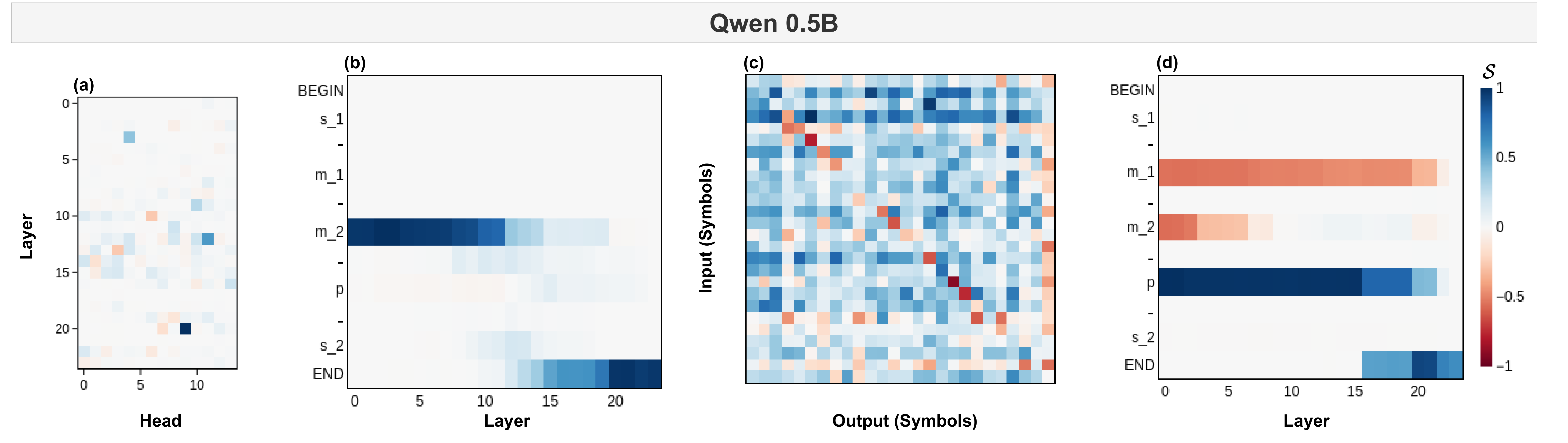}
\par\bigskip
\includegraphics[width=1.0\linewidth]{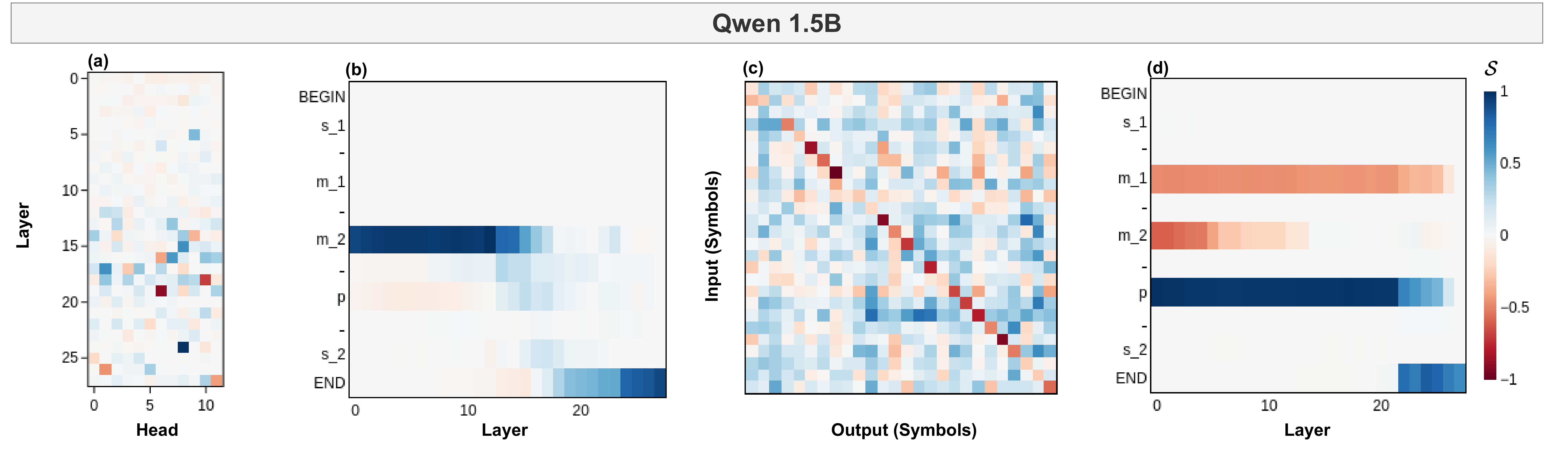}
\par\bigskip
\includegraphics[width=1.0\linewidth]{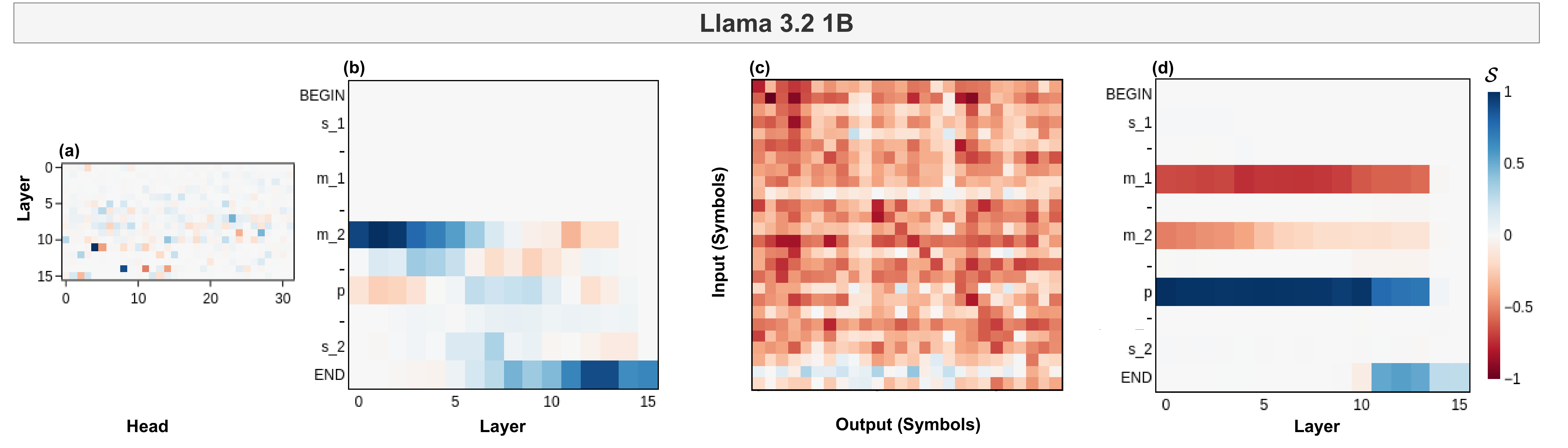}
\par\bigskip
\includegraphics[width=1.0\linewidth]{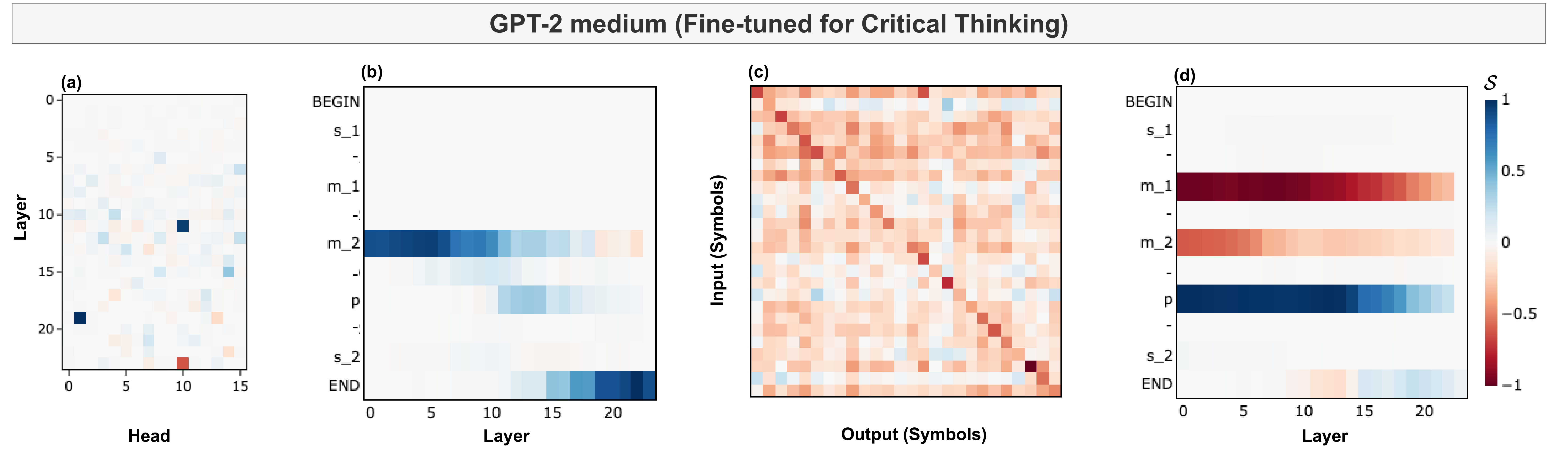}
\caption{Comprehensive results of the symbolic circuit analysis across different series of models (Qwen2.5, Llama3.2 and fine-tuned GPT-2 medium). (a) Attention output patching results and (b) residual stream patching results in the middle-term intervention setup. (c) OV circuit logit lens results for $m$-suppression head, with input and output comprising 26 uppercase letters. (d) Residual stream patching results in the all-term corruption setup. For clarity, a dash (–) indicates the averaged values for tokens appearing between terms.}
\label{fig:different_model_generalisability2}
\end{figure*}

Figures \ref{fig:different_model_generalisability} and \ref{fig:different_model_generalisability2} represent the results of applying the same circuit analysis in different model families. The $m$-suppression heads for each model were manually selected based on their likelihood, determined by patching scores and positioning within the activation pattern.

\end{document}